\pdfoutput=1
\documentclass{article}

\usepackage{microtype}
\usepackage{graphicx}
\usepackage{subfigure}
\usepackage{booktabs} 
\usepackage{caption}
\usepackage{subcaption}
\usepackage{url}
\usepackage{graphicx}

\usepackage{hyperref}

\usepackage[accepted]{icml2024}
\usepackage{amsmath}
\usepackage{amssymb}
\usepackage{mathtools}
\usepackage{amsthm}
\usepackage{multirow}

\usepackage[capitalize,noabbrev]{cleveref}

\theoremstyle{plain}

\theoremstyle{definition}

\theoremstyle{remark}

\usepackage[textsize=tiny]{todonotes}

\icmltitlerunning{Knowledge Transfer from Vision Foundation Models for Efficient Training of Small Task-specific Models}

\begin{document}

\twocolumn[

\icmltitle{Knowledge Transfer from Vision Foundation Models for \\Efficient Training of Small Task-specific Models }



\icmlsetsymbol{equal}{*}

\begin{icmlauthorlist}
\icmlauthor{Raviteja Vemulapalli}{comp}
\icmlauthor{Hadi Pouransari}{comp}
\icmlauthor{Fartash Faghri}{comp}
\icmlauthor{Sachin Mehta}{comp}
\icmlauthor{Mehrdad Farajtabar}{comp}
\icmlauthor{Mohammad Rastegari}{comp}
\icmlauthor{Oncel Tuzel}{comp}
\end{icmlauthorlist}

\icmlaffiliation{comp}{Apple, USA}

\icmlcorrespondingauthor{Raviteja Vemulapalli}{r$\_$vemulapalli@apple.com}
\icmlcorrespondingauthor{Hadi Pouransari}{mpouransari@apple.com}

\icmlkeywords{Efficient learning, Knowledge distillation, Foundation models}

\vskip 0.3in
]



\printAffiliationsAndNotice{} 

\begin{abstract}
Vision Foundation Models (VFMs) pretrained on massive datasets exhibit impressive performance on various downstream tasks, especially with limited labeled target data. However, due to their high inference compute cost, these models cannot be deployed for many real-world applications. Motivated by this, we ask the following important question, \textit{\lq\lq How can we leverage the knowledge from a large VFM to train a small task-specific model for a new target task with limited labeled training data?\rq\rq}, and propose a simple task-oriented knowledge transfer approach as a highly effective solution to this problem. Our experimental results on five target tasks show that the proposed approach outperforms task-agnostic VFM distillation, web-scale CLIP pretraining, supervised ImageNet pretraining, and self-supervised DINO pretraining by up to 11.6\%, 22.1\%, 13.7\%, and 29.8\%, respectively. Furthermore, the proposed approach also demonstrates up to $9\times, 4\times$ and $15\times$ reduction in pretraining compute cost when compared to task-agnostic VFM distillation, ImageNet pretraining and DINO pretraining, respectively, while outperforming them. We also show that the dataset used for transferring knowledge has a significant effect on the final target task performance, and introduce a retrieval-augmented knowledge transfer strategy that uses web-scale image retrieval to curate effective transfer sets.
\end{abstract}

\section{Introduction}
\label{sec:introduction}
\begin{figure}[!t]
    \centering
    \includegraphics[scale=0.96,bb=0 0 225 190]{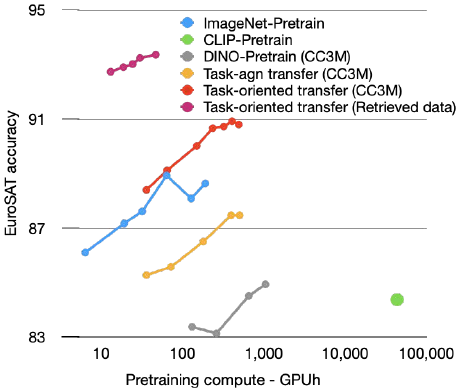}
    \caption{Downstream task performance (EuroSAT classification) of FastViT target model with different pretraining approaches. Here, we finetune the pretrained models using 10 labeled training images per class. Task-oriented knowledge transfer from DINOv2 VFM with generic web data (CC3M) outperforms the popular ImageNet, CLIP and DINO pretraining approaches. Knowledge transfer with a transfer set curated using image retrieval performs significantly better than knowledge transfer with generic web data.}
    \label{fig:pretraining_strategies}
\end{figure}

Currently, the computer vision community is witnessing the emergence of various vision and multi-modal foundation models pretrained on massive datasets~\citep{DINOV2, CLIP, Florence, BLIP2, Beit3}. These models have been shown to work well for many downstream tasks, especially, when task-specific labeled data is limited~\citep{CLIP}. However, they cannot be used for many resource-constrained applications due to their high inference computate cost. Also, many applications such as autonomous driving, medical image diagnostics, and industrial automation, focus on specific tasks or domains and need small task-specific models rather than a large Vision Foundation Model (VFM). This raises an important, timely and yet underexplored question: \textit{How can we leverage the knowledge from a large VFM to effectively train a small task-specific model for a new target task with limited labeled training data?}
\begin{figure*}[ht]
    \centering
    \includegraphics[scale=0.47,bb=0 0 1000 600]{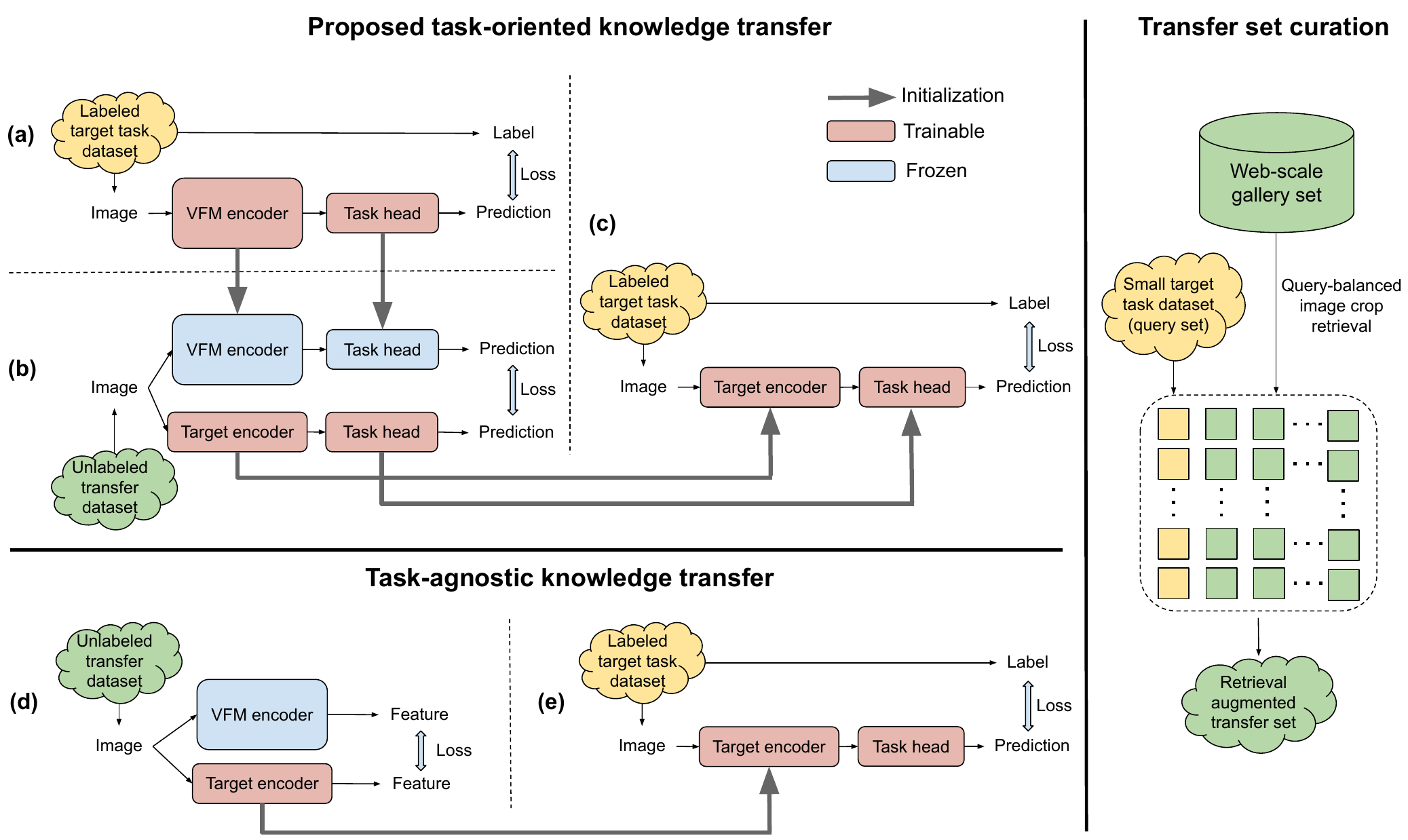}
    \caption{\textbf{Top-left:} Proposed task-oriented knowledge transfer approach that (a) first teaches the target task to the VFM using labeled target task data, (b) then uses this VFM to pretrain the target model by matching their target task predictions on an unlabeled transfer dataset, and (c) finally finetunes the target model using labeled target task data. \textbf{Bottom-left:} Alternative task-agnostic knowledge transfer approach that (d) first pretrains the target model by matching its features to the features extracted by the VFM on an unlabeled transfer dataset, and (e) then finetunes it using labeled target task data. \textbf{Right:} Transfer set curation using query-balanced image crop retrieval with a small target task dataset as the query set and a web-scale gallery set. By retrieving equal number of samples for each query, this approach increases the diversity of the retrieved samples. We perform crop-level retrieval to increase the chances of finding good matches.}
    \label{fig:knowledge_transfer}
\end{figure*}

Answering this question requires transferring knowledge from a VFM across both task and model architecture boundaries. This is different from the knowledge distillation setting that only focuses on knowledge transfer between model architectures~\citep{KD,Contrastive} and the transfer learning setting that only focuses on knowledge transfer between tasks~\citep{TLSurvey}.

\section{Approach and Contributions}\label{sec:approach}
In this work, we propose a simple and highly effective approach for transferring knowledge from a large pretrained VFM to a small task-specific model. This approach, referred to as \emph{\textbf{task-oriented knowledge transfer}}, first teaches the target task to the VFM using an appropriate task-specific head and limited labeled target task data, and then transfers task-oriented knowledge from the adapted VFM to the target model using the knowledge distillation framework of~\citep{KD} with a large unlabeled dataset, referred to as the \emph{transfer set}. Finally, the target model is finetuned with limited labeled target task data (see~\cref{fig:knowledge_transfer} Top-left).

An alternative approach to train a small task-specific model by leveraging a VFM is to first distill the VFM image encoder to the target model image encoder and then finetune the target model using limited labeled target task data (see~\cref{fig:knowledge_transfer} Bottom-left). We refer to this approach as \emph{\textbf{task-agnostic knowledge transfer}}. Both task-oriented and task-agnostic knowledge transfer approaches leverage VFMs that have been trained on web-scale datasets. Instead, one could  pretrain the small target model directly on a web-scale dataset. However, such pretraining can be extremely expensive.
For example, training a MobileViT-V2 model~\cite{MobilevitV2} using 0.7B image-text pairs following the CLIP approach of~\citep{CLIP} takes around 1.2K A100 GPU hours just for one epoch.

We compare the proposed task-oriented knowledge transfer approach with task-agnostic knowledge transfer from VFMs, web-scale CLIP pretraining, supervised ImageNet pretraining, and self-supervised DINO~\citep{DINO} pretraining on five target tasks under limited labeled data settings~\footnote{The target task data splits can be found at \url{https://github.com/apple/ml-vfm-kt/tree/main}.} using two VFMs, namely DINOv2-ViT-L/14~\citep{DINOV2} and OpenCLIP-ViT-L/14~\citep{Openclip}, and two target mobile architectures, namely FastViT-S12~\citep{Fastvit} and MobileViT-V2-1.0. We experiment with two transfer sets, a generic CC3M~\citep{CC3M} transfer set and a target task-related transfer set, and present several insightful findings to the community. To the best of our knowledge, there is no existing work that concretely establishes the below findings about leveraging VFMs for training small task-specific models with limited labeled data.

\begin{itemize}
    \item Task-oriented knowledge transfer outperforms task-agnostic transfer by a significant margin both in terms of performance (see~\cref{tab:improvement_results}) and training cost (see~\cref{fig:training_cost}). While VFMs can store vast knowledge due to their large capacity, small models may not be able to inherit this vast knowledge. Hence, transferring only task-oriented knowledge works better.

    \item Task-oriented knowledge transfer from VFMs outperforms the compute-intensive CLIP, widely-used supervised ImageNet and self-supervised DINO pretraining approaches by large margins (see~\cref{tab:improvement_results}), and is also computationally efficient (see~\cref{fig:pretraining_strategies,fig:training_cost}).

    \item While most of the existing distillation works use target task datasets as transfer sets, we systematically study the effect of transfer set. We show that task-oriented knowledge transfer outperforms ImageNet and CLIP pretraining  without using task-related transfer sets (e.g., by using a large-scale generic web dataset such as CC3M, see~\cref{tab:improvement_results}). The performance further improves when large task-related unlabeled datasets are used as transfer sets (see~\cref{fig:main_results}).

    \item When a large task-related transfer set is not readily available, we propose to curate such a dataset using web-scale image retrieval with the limited target task dataset as the query set (see~\cref{fig:knowledge_transfer}-Right). Retrieval-augmented transfer sets outperform the generic CC3M transfer set by a significant margin (see~\cref{tab:retrieval_results}).
\end{itemize}
\section{Experimental Analysis}
\label{sec:experiments}
\subsection{Experimental Setup}
We perform the target model training in two stages: \textit{pretraining} followed by \textit{finetuning}.  During pretraining, we utilize a VFM by following the task-oriented and task-agnostic knowledge transfer approaches presented in Sec. 2 using a large unlabeled dataset as the transfer set. During finetuning, we train the model on a small labeled target task dataset.

\textbf{Alternative approaches:} \emph{IN-Pretrain:} Supervised pretraining on 1.28M labeled images from the ImageNet-1K dataset~\citep{ImageNet}. \emph{CLIP-Pretrain:} Contrastive language image pretraining on 0.7B image-text pairs from~\citep{CLIPDataset} following the affinity mimicking CLIP distillation approach of~\citep{TinyCLIP}.~\footnote{~\citep{TinyCLIP} has shown that affinity mimicking distillation works better than the CLIP loss of~\citep{CLIP}. We use the best ViT-B/16 model from~\cite{RangeAugment} as teacher.} \emph{DINO-Pretrain:} Self-supervised pretraining on the unlabeled transfer set following the DINO approach of~\citep{DINO}. We finetune the target model on labeled target task data after IN/CLIP/DINO pretraining.

\textbf{Target tasks:} 
We present results on five diverse target tasks, namely HAM10K skin disease classification~\citep{Ham10k}, EuroSAT land cover classification~\citep{EuroSAT}, Places365 scene classification~\citep{Places365}, ImageNet object classification~\citep{ImageNet}, and ADE20K semantic segmentation~\citep{Ade20k}. See~\cref{sec:datasets} for details of the corresponding datasets. 

\textbf{Transfer sets}: For each target task, we experiment with two transfer sets. The first one is a generic transfer set consisting of 2.87M unlabeled images from the training split of the CC3M dataset~\citep{CC3M}, and the second one is a task-related transfer set consisting of unlabeled images from the target task domain. For each task, we use the entire training split of the corresponding dataset as the task-related transfer set. See~\cref{sec:transfer_sets} for further details. For ADE20K segmentation and EuroSAT classification tasks, we also experiment with transfer sets curated using image retrieval. See~\cref{sec:retrieval_details} for details of the retrieval process.

\textbf{Vision foundation models:} We use DINOv2-ViT-L/14 model~\citep{DINOV2} and the OpenCLIP-ViT-L/14 model~\citep{Openclip} trained on the DataComp-1B dataset~\citep{Datacomp} as VFMs. For brevity, we refer to them as DINOv2 and OpenCLIP, respectively.

\begin{table*}[ht]
    \begin{scriptsize}
    \begin{center}
    \begin{tabular}{|l|ccccc|ccccc|}
    \hline
        \multirow{2}{*}{Improvement over} & \multicolumn{5}{c}{When using generic CC3M transfer set} & \multicolumn{5}{|c|}{When using target task-related transfer set} \\\cline{2-11}
          & HAM10K & EuroSAT & Places365 & ImageNet & ADE20K & HAM10K & EuroSAT & Places365 & ImageNet & ADE20K \\\hline
         Task-agnostic transfer & 0.38 - 2.91 & 1.25 - 5.27 & 1.88 - 4.89 & 1.90 - \textbf{11.6} & 3.41 - 10.5 & 1.01 - 3.53 & 0.71 - 2.35 & 1.38 - 3.47 & 3.40 - \textbf{6.30} & 1.11 - 6.25\\\hline
        CLIP-Pretrain & 2.89 - 6.79 & 1.81 - 9.55 & 2.20 - 5.50& 2.30 - 11.2 & 6.52 - \textbf{20.9} & 5.93 - 11.4 & 4.24 - 15.5 & 3.38 - 7.72 & 5.30 - \textbf{22.1} & 5.45 - 19.2 \\\hline
         IN-Pretrain & 0.93 - 4.27 & 0.21 - 3.73 & 2.19 - 8.17 & - & 5.12 - \textbf{13.7} & 3.97 - 8.93 & 2.64 - 9.64 & 3.37 - 9.89 & - & 3.53 - \textbf{12.0} \\\hline
         DINO-Pretrain & 0.20 - 2.89 & 2.59 - 6.94 & 2.88 - 8.46 & 2.43 - \textbf{29.8} & 7.86 - 15.4 & 4.70 - 8.76 & 5.57 - 12.6 & 2.72 - 4.50 & 4.00 - \textbf{21.7} & 8.84 - 18.1\\\hline

    \end{tabular}
    \end{center}
    \end{scriptsize}
    \caption{Performance gain of task-oriented knowledge transfer from VFMs when compared to other training strategies. These values are the minimum and maximum gains observed across all VFM and target model combinations we have experimented with for each dataset. Highest performance gain over each competing approach is highlighted in bold.
    }
    \label{tab:improvement_results}
\end{table*}

\textbf{Target models:} We use two recent efficient architectures, namely FastViT-S12~\citep{Fastvit} and MobileViT-V2-1.0 \citep{MobilevitV2}, as image encoders for the target models. For brevity, we refer to them as FastViT and MViT-V2, respectively. We present the results for FastViT in this section and the results for MViT-V2 in~\cref{sec:MVIT_results}.

\textbf{Task-specific heads:} For classification tasks, the task-specific head is a linear classifier. The input to the classification layer is the final CLS token features for DINOv2 and OpenCLIP models, and the output of the final average pooling layer for MViT-V2 and FastViT models. For segmentation tasks, the task-specific head consists of a DeepLabV3 segmentation head~\citep{DeepLabV3} on top of the final spatial feature map and a spatial upsampling layer. As segmentation tasks requires high resolution features, we modify the image encoders such that their output spatial resolution is $1/8$ of the input image resolution. For DINOv2 and OpenCLIP models, we use a convolution stride of 8 instead of 14 for the patch embedding layer and resize the position embeddings accordingly. For MViT-V2 and FastViT models, we change the last two stride-2 convolutions to stride-1 convolutions.

\textbf{Adapting VFM to target task:} When teaching a target task to a VFM, we consider two options: 1) training only the task-specific head with frozen VFM and 2) finetuning the entire VFM. We compare both models in terms of their target task performance and use the best one for task-oriented knowledge transfer. In our experiments with ViT-L/14 based VFMs, finetuning performed best in all cases (except when only 1\% labeled data is available for ImageNet dataset).

\textbf{Loss functions:} For finetuning, we use cross entropy loss, and for matching task predictions, we use KL divergence. For segmentation tasks, these losses are applied at each pixel. The loss function used for matching features depends on the VFM. In the case of OpenCLIP model, we use contrastive loss~\citep{Contrastive}. Since DINOv2 is trained to produce good patch features along with global image features, we evaluate both these features. We use contrastive loss with global image features and cosine similarity loss with patch features. See~\cref{sec:training_details} for additional training details.

\begin{figure*}[ht!]
\centering
\includegraphics[scale=0.58, bb=210 0 500 975]{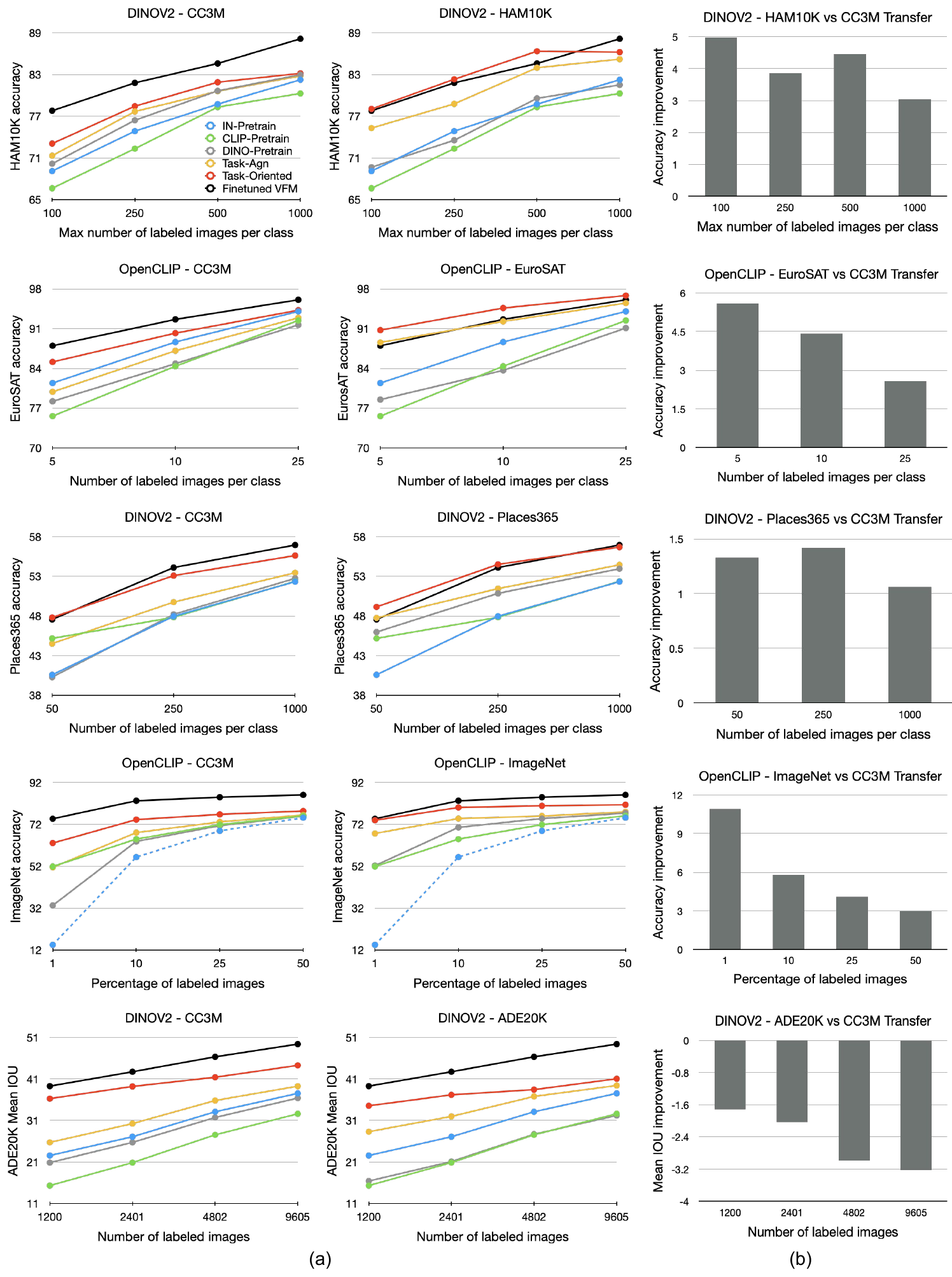}
\caption{(a) Comparison of various approaches for different (VFM, transfer set) combinations with FastViT as the target image encoder. (b) Performance improvement when unlabelled target task data is used instead of generic CC3M dataset for task-oriented knowledge transfer. The target tasks are HAM10K classification, EuroSAT classification, Places365 classification, ImageNet classification and ADE20K segmentation from top to bottom. Task-oriented knowledge transfer from VFMs (red curves) clearly outperforms alternative training strategies. The performance of finetuned VFMs used for knowledge transfer is also shown here for reference (black curves). When the target task is ImageNet classification, the blue curve corresponds to training from scratch instead of ImageNet pretraining.}
\label{fig:main_results}
\end{figure*}
\begin{figure*}[ht]
    \centering
    \includegraphics[scale=0.5, bb=520 0 400 220]{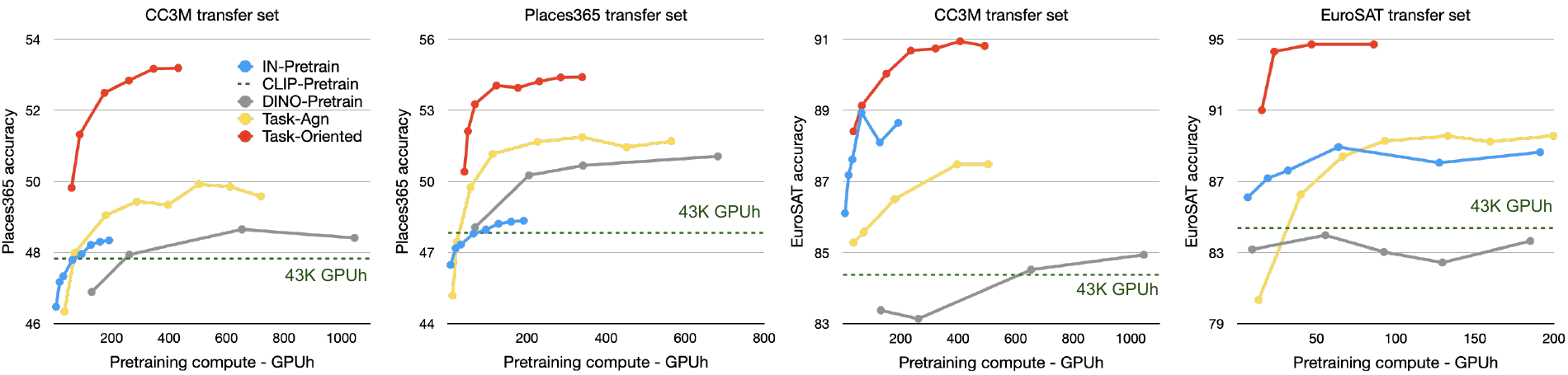}
    \caption{Comparison of various approaches in terms of their pretraining compute. The left two figures correspond to Places365 classification (250 training images per class) and the right two figures correspond to EuroSAT classification (10 training images per class). Here, we use DINOv2 VFM for knowledge transfer and FastViT as the target architecture. Each curve in this figure was obtained by evaluating intermediate checkpoints of one training run. CLIP pretraining is represented using dashed green line.}
    \label{fig:training_cost}
\end{figure*}

\subsection{Effectiveness of Task-oriented Knowledge Transfer}
\subsubsection{Performance Improvement}
\label{sec:performance_improvement}
\Cref{fig:main_results}(a) compares the performance of various approaches for different VFM and transfer set combinations on five downstream tasks with FastViT as the target image encoder. Please see~\Cref{sec:additional_fastvit_results} for additional results corresponding to FastViT model, and~\Cref{sec:MVIT_results} for results corresponding to MViT-V2 model. For task-agnostic knowledge transfer from DINOv2, we experiment with both global image features and patch features and report the best results. 

The proposed task-oriented knowledge transfer from VFMs outperforms task-agnostic transfer from VFMs, ImageNet, CLIP and DINO pretraining approaches for all target tasks irrespective of whether the generic CC3M transfer set or a target task-related transfer set is used.~\cref{tab:improvement_results} shows the minimum and maximum performance gains of task-oriented knowledge transfer when compared to other pretraining strategies. Specifically, when using the generic CC3M transfer set, the performance gains are up to 11.6\%, 20.9\%, 13.7\% and 29.8\% when compared to task-agnostic transfer, CLIP, ImageNet, and DINO pretraining approaches, respectively. When using target task-related transfer sets, the performance gains are up to 6.3\%, 22.1\%, 12\% and 21.7\% when compared to task-agnostic transfer, CLIP, ImageNet and DINO pretraining approaches, respectively.

When task-related transfer set is used, task-oriented transfer even outperforms finetuned VFM for some of the tasks when the amount of labeled data is small (see red and black curves in~\cref{fig:main_results})). This could be because the knowledge transfer process uses large-scale unlabeled target domain data that is not used while finetuning VFMs. 

\subsubsection{Training Efficiency}
\label{sec:training_cost}
The main focus of this work is on improving the performance of the target model under limited labeled data settings. Hence, for each competing approach, we train the target model long enough to achieve the best possible performance with the given limited labeled data.~\Cref{fig:training_cost} compares various approaches in terms of their pretraining compute cost measured using A100 GPU hours (GPUh) with FastViT as the target model. Task-oriented knowledge transfer from VFMs achieves better performance with significantly less training compute when compared to other pretraining strategies. Specifically, it demonstrates up to $9\times, 4\times$ and $15\times$ reduction in training compute when compared to task-agnostic knowledge transfer, ImageNet pretraining and DINO pretraining, respectively, while outperforming them. Web-scale CLIP pretraining of the target model performs significantly worse than task-oriented knowledge transfer while using significantly more compute (43K GPUh). 

Note that the training compute for task-oriented knowledge transfer includes the compute used for adapting the VFM to the target task. Since VFMs are trained only on small labeled datasets, the compute cost of adapting them is significantly lower than the compute cost of distilling them on large unlabeled datasets. For example, while finetuning DINOv2 VFM on Places365 dataset using 250 images per class for 100 epochs takes 32 GPUh, distilling the finetuned VFM on Places365 transfer set for 100 epochs takes 180 GPUh. Similarly, while finetuning DINOv2 VFM on EuroSAT dataset using 10 images per class for 200 epochs takes 4 GPUh, distilling the finetuned VFM on EuroSAT transfer set for 5K epochs takes 40 GPUh~\footnote{Close to best target task performance is observed after distilling finetuned DINOv2 for about 100 epochs on Places365 transfer set and 5K epochs on EuroSAT transfer set.}.

\begin{table}[t!]
    \begin{scriptsize}
    \begin{center}
    \begin{tabular}{|l|c|c|c|c|}
    \hline
        Dataset & HAM10K & EuroSAT & Places365 & ImageNet\\\hline
        Labeled data & 100 img/cls & 10 img/cls & 250 img/cls & 50\%\\\hline
        VFM & DINOv2 & OpenCLIP & OpenCLIP & OpenCLIP \\\hline
        \multicolumn{5}{c}{VFM performance}
        \\\hline
        VFM-LP & 71.10 & 87.33 & 52.39 & 84.07\\\hline
        VFM-FT & 76.92 & 92.63 & 54.48 & 86.05\\\hline
        \multicolumn{5}{c}{Task-oriented knowledge transfer with generic CC3M transfer set}\\\hline
        VFM-LP & 69.97 & 89.59 & 52.87  & 78.13 \\\hline
        VFM-FT & 73.06 & 90.21 & 53.33 & 78.40\\\hline
        \multicolumn{5}{c}{Task-oriented knowledge transfer with target task-related transfer set}\\\hline
        VFM-LP & 74.60 & 94.37 & 54.69 & 81.01 \\\hline
        VFM-FT & 78.04 & 94.63 & 54.82 & 81.43\\\hline
    \end{tabular}
    \end{center}
    \end{scriptsize}
    \caption{Comparison between linear-probed (LP) and finetuned (FT) VFMs in terms of their effectiveness for task-oriented knowledge transfer to FastViT target model.}
    \label{tab:vfm_finetune_vs_linearprobe}
    \vspace{-5pt}
\end{table}

\subsubsection{VFM Finetuning vs Linear Probing}
We consider both linear probing and full finetuning when adapting a VFM to a target task, and use the best-performing VFM for task-oriented knowledge transfer. In almost all of our experiments, finetuning resulted in a better performing VFM when compared to linear probing. In this section, we compare linear-probed and finetuned VFMs in terms of their effectiveness for task-oriented knowledge transfer to FastViT target model (see~\Cref{tab:vfm_finetune_vs_linearprobe}). While transferring knowledge from finetuned VFM outperforms transferring from linear-probed VFM by a significant margin in the case of HAM10K dataset $(> 3\%)$, the performance gap is small for other datasets $(< 0.5\%)$. It is interesting to see that the performance gap between target models distilled from linear-probed and finetuned VFMs is significantly smaller than the performance gap between the linear-probed and finetuned VFMs themselves. This suggests that \emph{full finetuning may not be needed for effective task-oriented knowledge transfer, especially as VFMs become more powerful}.

Note that the small target model exhibits better performance than the teacher VFM on many datasets when a task-related transfer set is used. This is potentially because the small model has been exposed to more task-related samples (additional unlabeled samples) when compared to teacher VFM.

\begin{table}[t!]
    \begin{scriptsize}
    \begin{center}
    \begin{tabular}{|l|l|c|c|}
    \multicolumn{4}{c}{Pretraining dataset - CC3M transfer set}\\
    \hline
      \multirow{2}{*}{ADE20K} & Number of labeled training images & 1200 & 2401\\\cline{2-4}
     &  Performance gain & 2.19 & 2.03 \\\hline\hline
   \multirow{2}{*}{ImageNet} & Percentage of labeled training images & 10 & 25\\\cline{2-4}
      &  Performance gain & 3.8 & 5.8\\\hline\hline
    \multirow{2}{*}{HAM10K} & Maximum labeled training images per class & 250 & 500\\\cline{2-4}
    & Performance gain & 9.99 & 7.41\\\hline\hline  
    \multirow{2}{*}{EuroSAT} & Number of labeled training images per class & 5 & 10\\\cline{2-4}
    & Performance gain & 3.3 & 4.9\\\hline\hline
    \multirow{2}{*}{Places365} & Number of labeled training images per class & 250 & 1000\\\cline{2-4}
    & Performance gain & 0.47 & 0.76\\\hline
    \multicolumn{4}{c}{Pretraining dataset - Target task-related transfer set}\\\hline
     \multirow{2}{*}{ADE20K} & Number of labeled training images & 1200 & 2401\\\cline{2-4}
     &  Performance gain & 1.23 & 1.25 \\\hline
    \end{tabular}
    \end{center}
    \end{scriptsize}
    \caption{Performance gain (for FastViT model) due to the final finetuning step of task-oriented knowledge transfer approach.}
    \label{tab:contribution_of_finetuning}
    \vspace{-5pt}
\end{table}

\subsubsection{Contributions of Finetuning Step}
The last step of the proposed task-oriented knowledge transfer approach is to finetune the target model on limited labeled data (see~\Cref{fig:knowledge_transfer}(c)). One may wonder to what extent this finetuning step contributes to the performance, given that the target model learns to follow a VFM finetuned on the target task in the pretraining step.~\Cref{tab:contribution_of_finetuning} shows the performance improvements due to the finetuning step. We see significant gains when generic CC3M transfer set is used during pretraining. When target task-related transfer set is used during pretraining, we see significant gains in the case of ADE20K segmentation task. We did not observe noticeable gains for the classification tasks in this case.

\subsection{Importance of Transfer Set}
\label{sec:transferset_importance}

\Cref{fig:main_results}(b) shows the improvement in accuracy on five downstream tasks when target task-related transfer sets are used instead of generic CC3M transfer set for task-oriented knowledge transfer from various VFMs to FastViT target model. Please see~\Cref{sec:additional_fastvit_results} for additional results with FastViT target model and~\Cref{sec:MVIT_results} for results corresponding to MViT-V2 target model.

Using task-related transfer sets improves the accuracy by up to 5\%, 7.3\%, 2.3\% and 10.9\% for HAM10K, EuroSAT, Places365 and ImageNet classification tasks, respectively (including the results presented in~\Cref{sec:additional_fastvit_results,sec:MVIT_results}). This underscores the importance of the relevance of transfer set to the target task. However, for ADE20K segmentation task, the generic CC3M transfer set performs better. We conjecture that the main reason for this is the size of ADE20K transfer set which has only 19.2K images. We address this issue by curating a larger task-related transfer set using image retrieval as shown in the next section.

\subsection{Retrieval Augmented Knowledge Transfer} \label{sec:rats}
Our results on HAM10K, EuroSAT, Places365 and ImageNet classification tasks show that using a sufficiently large target task-related transfer set performs better than a generic transfer set such as CC3M. However, such large transfer sets may not be readily available for every task. We propose to address this issue by curating large task-related transfer sets using image retrieval as shown in \cref{fig:knowledge_transfer}-Right.

Specifically, we use the limited target task labeled dataset as the query set $\mathcal{Q}$ and a large set of images sourced from web as the gallery $\mathcal{G}$. We use an encoder network $\phi$ to map all the images to a $d$-dimensional embedding space, and perform retrieval based on Euclidean distances in this space following a query balanced approach. For a query image $x_q \in \mathcal{Q}$, we define $k$-NN$(x_q)$ to be the set of its $k$ nearest neighbors from $\mathcal{G}$. To retrieve $N$ images in total, we find the smallest $k$ for which $\bigcup_{x_q \in \mathcal{Q}}k$-NN$(x_q)$ contains at least $N$ unique images. If $\bigcup_{x_q \in \mathcal{Q}}k$-NN$(x_q)$ contains more than $N$ images, we drop the $k^{th}$ neighbor of randomly selected queries until the retrieved set has $N$ images.  By giving equal weight to all the queries, this approach increases the diversity of the retrieved samples. We combine the retrieved set with the initial query set and use this retrieval augmented transfer set for task-oriented knowledge transfer.

\begin{figure}[t!]
\centering
\includegraphics[scale=0.4,bb=290 0 290 250]{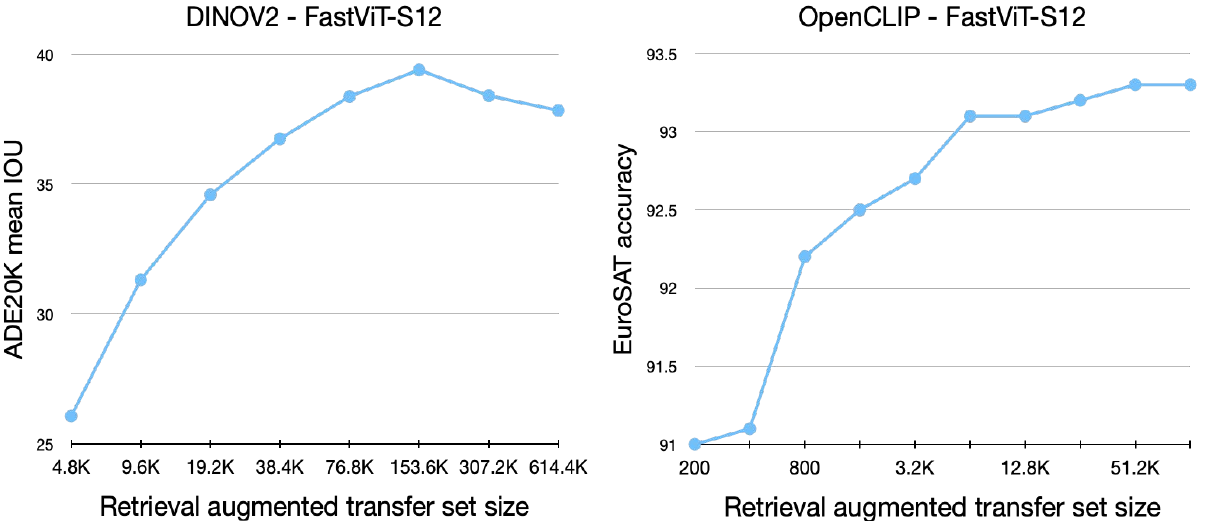}
\caption{Performance of task-oriented knowledge transfer using retrieval augmented transfer sets of varying sizes. The number of labeled images used for finetuning and also as retrieval queries is 4800 for ADE20K dataset and 100 for EuroSAT dataset.}
\label{fig:transferset_size_ablation}
\vspace{-5pt}
\end{figure}

\begin{figure}[!t]
    \centering
    \includegraphics[scale=0.32, bb=0 10 400 450]{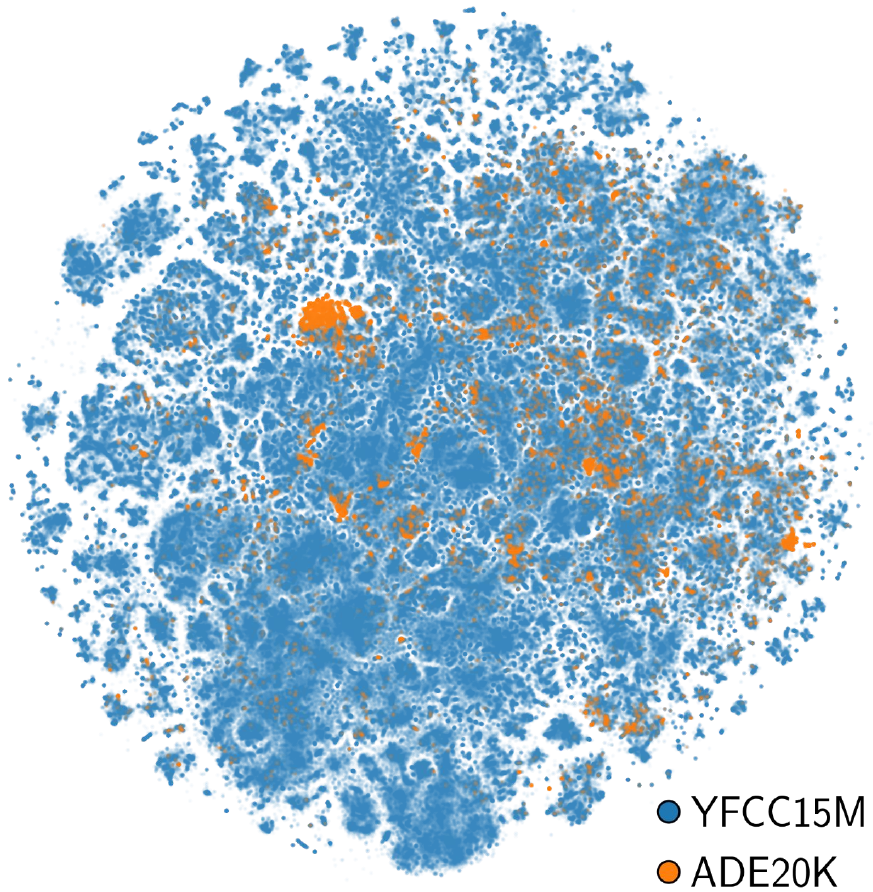}
    \caption{t-SNE~\citep{TSNE} visualization of image features of ADE20K dataset and randomly sampled 10\% of YFCC15M dataset.}
    \label{fig:tsne-yfcc}
\vspace{-5pt}    
\end{figure}

\subsubsection{ADE20K Segmentation}
We use YFCC15M dataset~\cite{CLIP} which contains 15M images as the gallery set, OpenCLIP-ViT-L/14 image encoder~\citep{Openclip} trained on the DataComp-1B dataset~\citep{Datacomp} as the encoder network, and experiment with the combination of DINOv2 VFM and FastViT target model. See~\cref{sec:image-level-ret} for details of the retrieval process and~\cref{fig:ade20_yfcc15m} in~\cref{sec:ret_visual} for some examples of retrieval results.

\Cref{fig:transferset_size_ablation} (left) shows ADE20K segmentation performance of task-oriented knowledge transfer with retrieval augmented transfer sets of different sizes. Here, we use 4800 labeled images as both finetuning dataset and retrieval query set. The segmentation performance increases with the transfer set size until we reach 154K images and drops after that. \Cref{fig:tsne-yfcc} displays a t-SNE visualization of image features from the ADE20K dataset and a randomly sampled 10\% of the YFCC15M dataset. Images from the ADE20K dataset occupy only a small region suggesting that only a small subset of YFCC15M is relevant for ADE20K segmentation. Hence, as we increase the size of retrieved dataset, after certain point, low quality matches will be added to the transfer set resulting in performance drop.
\begin{table}[!t]
    \begin{scriptsize}
    \begin{center}
    \begin{tabular}{|l|cccc|}
         \multicolumn{5}{c}{ADE20K mean IOU (transfer from DINOv2 VFM)}\\\hline
         Labeled images / Query set size & 1200 & 2401 & 4802 & 9605 \\\hline
         Full ADE20K transfer (19.2K) & 34.57 & 37.19 & 38.45 & 41.07\\\hline 
         CC3M transfer (2.87M) & 36.28 & 39.22 & 41.44 & 44.29 \\\hline
         Retrieval augmented transfer (154K) & 37.65 & 40.40 & 43.28 & 44.93\\\hline
         \end{tabular}
         \begin{tabular}{|l|ccc|}
        \multicolumn{4}{c}{EuroSAT accuracy (transfer from OpenCLIP VFM)}\\\hline
        Labeled images / Query set size & 50 & 100 & 250\\\hline
        Full EuroSAT transfer (2.7K) & 90.74 & 94.63 & 96.83 \\\hline 
        CC3M transfer (2.87M) & 85.14 & 90.21 & 94.25 \\\hline
        Retrieval augmented transfer (51K) & 89.23 & 93.25 & 96.35 \\\hline
    \end{tabular}
    \end{center}
    \end{scriptsize}
    \vspace{-5pt}    
    \caption{Comparison between various transfer sets in terms of their effectiveness for task-oriented knowledge transfer.}
    \label{tab:retrieval_results}
    \vspace{-10pt}    
\end{table}

Using 154K as the target transfer set size, we curate different transfer sets by varying the number of query images used for retrieval. \Cref{tab:retrieval_results} (top) compares these curated transfer sets with the generic CC3M and the full ADE20K transfer sets. Retrieval augmented transfer sets clearly outperform both ADE20K and generic CC3M transfer sets.~\footnote{The best performance in~\cref{fig:transferset_size_ablation}(left) is lower than the performance for curated transfer set corresponding to 4800 query images in~\cref{tab:retrieval_results}(top). This is because, we used shorter training runs (60K steps) to get the results in~\cref{fig:transferset_size_ablation}(left), and once we identified the best transfer set size, we used longer training runs (200K steps) to get the results in~\cref{tab:retrieval_results}(top).}


\subsubsection{EuroSAT Classification}
\begin{figure}[!t]
    \centering
    \includegraphics[scale=1.05, bb=200 0 15 215]{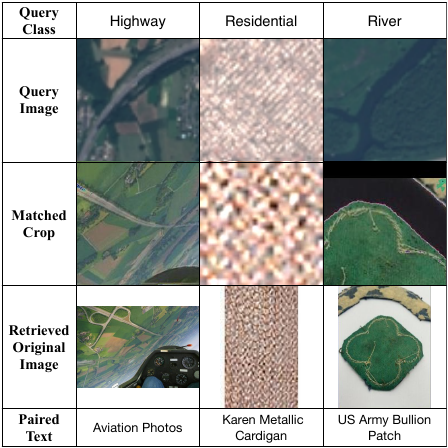}
    \caption{Images retrieved from DataComp-1B dataset for queries from the EuroSAT dataset. Proposed cropping and augmentation-based retrieval strategy selects only the relevant regions from gallery images (see the Highway example). It also enables the identification of regions with similar patterns from out-of-domain images (see the Residential and River examples).}
    \label{fig:retrieval_results}
\end{figure}

Since this dataset contains specialized imagery (satellite images), we use a much larger gallery to increase the chances of finding good matches. Specifically, we use the DataComp-1B dataset~\cite{Datacomp} which contains 1.28B images filtered from Common Crawl (web-crawled data). We take 10 randomly augmented crops from each image and create a gallery of 12.8B images. See~\cref{sec:crop-level-ret} for additional details about this gallery set and the retrieval process. We show some examples of retrieval results in~\cref{fig:retrieval_results}. Our crop-level retrieval strategy enables the extraction of relevant regions from  gallery images that are both in-domain and out-of-domain relative to the query. See~\cref{fig:eurosat_datacomp} in~\cref{sec:ret_visual} for additional examples.

\Cref{fig:transferset_size_ablation}(right) shows EuroSAT classification performance using task-oriented knowledge transfer with retrieval augmented transfer sets of different sizes. Here, we use a dataset of 100 labeled images (10 images per class) both as finetuning dataset and retrieval query set. The performance saturates at about 51K. Using 51K as the target size, we curate different transfer sets by varying the number of query images used for retrieval.~\Cref{tab:retrieval_results} (bottom) compares these curated transfer sets with the generic CC3M and the full EuroSAT transfer sets. Retrieval augmented transfer sets curated using small query sets outperform the generic CC3M transfer set by a significant margin, and perform competitively when compared to the full EuroSAT transfer set.
\section{Related Works}
\textbf{Knowledge distillation} is a widely-used approach for transferring knowledge between model architectures by training one model to mimic the outputs of another model. Numerous distillation approaches have been proposed over the past decade based on various knowledge representations such as task logits~\citep{KD}, intermediate features or embeddings~\citep{FeatureDistill,Contrastive}, relations between samples~\citep{RelationsDistill}, attention maps~\citep{AttentionDistill}, etc. Please refer to~\citet{KDSurvey,KDSurvey1} for an overview of existing distillation approaches. Some recent works have specifically focused on multi-modal distillation of image-language models~\citep{VLM-KD, VLKD, EfficientVLM, DIME-FM,CLIP-KD,MobileCLIP}. In addition to transferring knowledge between model architectures, this work also focuses on transferring knowledge between tasks.

\textbf{Transfer learning}, where a model is first pretrained on a data-rich task before being partially or fully finetuned on a downstream task, has been well studied over the past decade~\citep{TLSurvey1, TLSurvey}. 
Recently,~\citep{PretrainData} compared various pretraining approaches and showed that supervised ImageNet training and large-scale CLIP training are effective pretraining strategies for several downstream vision tasks. While the standard transfer learning setting focuses on transferring knowledge only between tasks, this work focuses on transferring knowledge between both tasks and model architectures.

\textbf{Image retrieval} has been used by various recent works to curate training datasets~\citep{SuSX,IE,CIT,NeuralPrime,REACT}. While~\citep{IE} focuses on self-supervised learning, the remaining works focus on training or adapting vision-language models. Different from these works, we use retrieval to curate task-related datasets for transferring knowledge from VFMs to small task-specific models.

\textbf{Self-supervised learning}, which uses unlabeled data to obtain a good initial feature representation, has received significant attention in the recent past, and several approaches have been proposed based on contrastive learning~\citep{SimCLR,MoCo}, distillation~\citep{BYOL,SimSiam,DINO}, redundancy reduction~\citep{BarlowTwins}, clustering~\citep{DeepCluster,SWAV} and image inpainting~\citep{MAE,Beit}. Please refer to~\citet{SSLSurvey} for a detailed review of existing self-supervised learning approaches.

\textbf{Semi-supervised learning} approaches leverage both labeled and unlabeled data to improve the final task performance. They focus on effectively propagating label information from a labeled dataset to an unlabeled dataset~\citep{PsuedoLabel,NoisyStudent}, and training the network using consistency constraints on the unlabeled samples~\citep{MeanTeacher,MixMatch,UDA,FixMatch,ICT}. Please refer to~\citep{SemiSupSurvey} for a recent survey of semi-supervised approaches. 


\textbf{Task-agnostic knowledge transfer from VFMs} has been explored in recent works such as DINOv2~\cite{DINOV2} and GSD~\cite{GSD}. However, they did not evaluate the effectiveness of task-agnostic transfer in limited labeled data settings. Also, they use distillation approaches specifically designed for transformer architectures.

\textbf{Task-oriented knowledge transfer} has been recently explored in the context of large langugae models (LLMs) by~\citep{LLMDistill,LLMDistill1}. These approaches use the rationales extracted from LLMs by chain-of-thought prompting to train small task-specific models. In this work, we focus on vision foundation models.

Recently,~\cite{DistillNearest} has also focused on training small task-specific models with limited labeled data. They use a pool of small CNN models trained on small-scale classification datasets as source models and focus on finding a set of suitable source models to distill for a given target task. Different from this work, we focus on knowledge transfer from recent VFMs that have been trained on massive datasets, and show that transferring knowledge from a single VFM outperforms various popular pretraining strategies. While~\cite{DistillNearest} assumes the existence of large-scale unlabeled target domain data, we show that knowledge transfer from VFMs is effective even when using generic web data such as CC3M. We also propose a retrieval-based approach for curating effective task-related transfer sets.

\section{Conclusions}
In this work, we proposed a simple yet highly effective task-oriented knowledge transfer approach for training small task-specific models by leveraging pretrained VFMs. 
Our experimental results on five target tasks show that the proposed approach outperforms task-agnostic VFM distillation, web-scale CLIP pretraining, supervised ImageNet pretraining, and self-supervised DINO pretraining approaches by a significant margin both in terms of target task performance and training efficiency. We showed that while transferring knowledge from VFMs using task-related transfer sets works best, using general web data such as CC3M is also highly effective when compared to popular ImageNet, CLIP and DINO pretraining approaches. We also proposed a crop-based retrieval strategy to curate large task-related transfer sets, and experimentally demonstrated its effectiveness. In summary, this work advocates for \textit{pretraining small task-specific models by transferring task-oriented knowledge from VFMs using large task-related transfer sets when they are available and retrieval-augmented transfer sets in the absence of large task-related transfer sets.}

\textbf{Future work:} In this work, we only used labeled target task data to finetune VFMs. A potential future work is to leverage additional large-scale unlabeled data to better adapt VFMs to the target task/domain, thereby eventually improving the small task-specific model trained with knowledge transfer from VFMs. However, finetuning VFMs on large scale datasets could be computationally expensive. Another potential future work is to explore more sophisticated transfer set curation strategies based on active learning and adversarial filtering that may result in better transfer sets that lead to better target model performance.

\textbf{Limitations:} Though our crop-based retrieval strategy enables fine-grained retrieval from a diverse set of images, curating large transfer sets could still be difficult for some specialized domains such as industrial automation that are not well covered by web data.
\section*{Acknowledgements} We thank Ting-Yao Hu, Karren Yang, Mohammad Samragh Razlighi, Pavan Kumar Anasosalu Vasu, and Barry Theobald from Apple for their valuable suggestions.
\bibliography{main}
\bibliographystyle{icml2024}
\newpage
\appendix
\onecolumn
\section{Experimental Setup}
\subsection{Target Task Datasets}
\label{sec:datasets}
\textbf{Places365 scene classification}~\citep{Places365}: This dataset has 1.8M training and 36.5K validation images. We split the original validation set into two subsets consisting of 3.65K and 32.85K images, and use them for validation and testing, respectively. This is a 365-class classification task.

\textbf{HAM10K skin lesion disease classification}~\citep{Ham10k}: This dataset consists of 10K training, 193 validation and 1.5K test images. HAM10K dataset is highly imbalanced with just 115 training images in the smallest class and 6705 training images in the largest class. When experimenting with $N$ training images per class, if a class does not have $N$ images, we just use all the images from that class. This is a 7-class classification task.

\textbf{EuroSAT land cover classification}~\cite{EuroSAT}: This dataset consists of 27K images. We follow the most difficult setting in~\citep{EuroSAT} and use 10\% of the dataset (2.7K images) as training split. We split the remaining 90\% dataset into two splits consisting of 5.4K and 18.9K images, and use them for validation and testing, respectively. This is a 10-class classification task.

\textbf{ADE20K semantic segmentation}~\citep{Ade20k}: This dataset consists of 20.2K training and 2K validation images. We split the original training set into two subsets with 19.2K and 1K images, and use them for training and validation, respectively. We use the original 2K validation set as the test set. This is a semantic segmentation task with 150 semantic classes.

\textbf{ImageNet}~\cite{ImageNet}: This dataset consists of 1.28M training images and 50K validation images. We report final accuracy results on the validation split. This is a 1000-class classification task.

The training, validation and test splits used in this work can be found at \url{https://github.com/apple/ml-vfm-kt/tree/main}.

\subsection{Transfer Sets}
\label{sec:transfer_sets}
The generic CC3M transfer set consists of 2.87M images. For each target task, we use the entire training split of the corresponding dataset as the task-related transfer set. So, the Places365, ImageNet, AD20K, HAM10K and EuroSAT transfer sets consist of 1.8M, 1.28M, 19.2K, 10K and 2.7K images, respectively.

\subsection{Training Details}
\label{sec:training_details}
We use AdamW optimizer~\citep{AdamW} with cosine learning rate decay in all our experiments. We use input resolutions of $256\times 256$ and $512\times 512$ for classification and segmentation tasks, respectively. 

\textbf{Loss functions:}
The loss function used for matching features depends on the VFM. In the case of OpenCLIP model, we use contrastive loss~\citep{Contrastive} with a linear projection layer on top of the target model output to match its dimensionality with the CLIP embedding dimensionality ($d=768$). In the case of DINOv2 global image features, we use contrastive loss with a linear projection layer ($d=768$) on outputs of both DINOv2 and target models. In the case of DINOv2 patch features, we use cosine similarity loss with a linear projection layer on top of the target model features to match their dimensionality with DINOv2 feature dimensionality ($d = 1024$). We also resize DINOv2 patch features to match the spatial resolution of the target model features.

\textbf{Augmentations and resolution:}
We adopt the advanced image augmentations used for ImageNet-1k in~\citep{MobilevitV2} and an input resolution of $256\times 256$ for ImageNet pretraining, task-agnostic VFM distillation, VFM classification finetuning, classification task logits distillation, and target classification model finetuning. We adopt the segmentation task-related augmentations used in~\citep{MobilevitV2} and an input resolution of $512\times 512$ for VFM segmentation finetuning, segmentation task logits distillation and target segmentation model finetuning. We use the augmentations from~\citep{DINO} for DINO pretraining and the augmentations from~\citep{CLIP} for CLIP pretraining with an input resolution of $256\times 256$.

\textbf{VFM finetuning:} When finetuning on Places365 dataset, we use a batch size of 512. We train for 200 epochs when 50 labeled images per class are used and 100 epochs when 250/1000 labeled images per class are used. When finetuning on HAM10K dataset, we use a batch size of 128 and train for 200 epochs. When finetuning on EuroSAT dataset, we train for 500 epochs with a batch size of 50 when using 5 labeled images per class, 400 epochs with a batch size of 100 when using 10 labeled images per class, and 300 epochs with a batch size of 128 when using 25 labeled images per class. When finetuning on ImageNet dataset, we use a batch size of 512 and train for 25 epochs when using 10\%, 25\% and 50\% labeled data. When using 1\% labeled data, we only train the linear classifier head for 100 epochs with a batch size of 1024. When finetuning on ADE20K dataset, we use a batch size of 32 and train for 300/250/200/100 epochs when using 1200/2401/4802/9605 labeled images. We run each finetuning experiment with several learning rates from $7e^{-6}$ to $3e^{-3}$ and report results corresponding to best ones.

\textbf{ImageNet pretraining}: We use a learning rate of $2e^{-3}$ and train with a batch size of 1024 for 300 epochs.

\textbf{CLIP pretraining:} We train for 200K iterations on 0.7B image-text pairs from~\citep{CLIPDataset} using a learning rate of $5e^{-4}$ and a batch size of 65.5K.

\textbf{DINO pretraining:} We use a learning rate of $3e^{-3}$ and train for 100, 200, 300, 10K, 15K and 20K epochs on CC3M, Places, ImageNet, EuroSAT, ADE20K and HAM10K datasets, respectively, with a batch size of 1024.

\textbf{Task-agnostic VFM feature distillation:} We train for 100 epochs with a batch size of 2048 when using CC3M transfer set. We train for 200 epochs with a batch size of 2048 when using Places365 and ImageNet transfer sets. We train for 10K and 20K epochs with a batch size of 1024 when using ADE20K and HAM10K transfer sets, respectively. When using EuroSAT transfer set, we train for 10K epochs using a batch size of 2048. We use a learning rate $1e^{-3}$ for these distillation experiments.

\textbf{Distillation of finetuned VFM:} When using CC3M transfer set, we train for 100 epochs with a learning rate of $7e^{-4}$ and batch size of 512 when distilling VFMs finetuned for Places365 and HAM10K classification tasks, 100 epochs with a learning rate of $7e^{-4}$ and a batch size of 1024 when distilling VFMs finetuned for EuroSAT classification task, 150 epochs with a learning rate of $3e^{-3}$ and a batch size of 1024 when distilling VFMs finetuned for ImageNet classification task, 15 epochs with a learning rate of $7e^{-4}$ and a batch size of 128 when distilling VFMs finetuned for ADE20K segmentation task. When using task-related transfer sets, we train for 200 epochs with a learning rate of $7e^{-4}$ and batch size of 512 when distilling VFMs finetuned for Places365 classification task, 7K epochs with a learning rate of $7e^{-4}$ and a batch size of 128 when distilling VFMs finetuned for HAM10K classification task, 10K epochs with a learning rate of $7e^{-4}$ and a batch size of 1024 when distilling VFMs finetuned for EuroSAT classification task, 300 epochs with a learning rate of $3e^{-3}$ and a batch size of 1024 when distilling VFMs finetuned for ImageNet classification task, 2K epochs with a learning rate of $7e^{-4}$ and a batch size of 128 when distilling VFMs finetuned for ADE20K segmentation task. When using curated transfer sets, we train for 550K steps with a learning rate of $7e^{-4}$ and a batch size of 128 when distilling VFMs finetuned for HAM10K classification task, 200K steps with a learning rate of $7e^{-4}$ and a batch size of 128 when distilling VFMs finetuned for ADE20K segmentation task, and 30K steps with a learning rate of $7e^{-4}$ and a batch size of 1024 when distilling VFMs finetuned for EuroSAT classification task. 

\textbf{Target model finetuning:} When finetuning on Places365 dataset, we use a batch size of 512. We train for 200 epochs when 50 labeled images per class are used and 100 epochs when 250/1000 labeled images per class are used. When finetuning on HAM10K dataset, we use a batch size of 128 and train for 200 epochs. When finetuning on EuroSAT dataset, we train for 800 epochs with a batch size of 50 when using 5 labeled images per class, 600 epochs with a batch size of 100 when using 10 labeled images per class, and 400 epochs with a batch size of 128 when using 25 labeled images per class. When finetuning on ImageNet dataset, we use a batch size of 512 and train for 50 epochs. When finetuning on ADE20K dataset, we use a batch size of 32 and train for 500/400/300/200 epochs when using 1200/2401/4802/9605 labeled images. We run each finetuning experiment with several learning rates from $7e^{-6}$ to $3e^{-3}$ and report results corresponding to best ones.

\section{Image Retrieval Details}
\label{sec:retrieval_details}
In this section, we describe the retrieval process used for curating task-related transfer sets. Given a query set $\mathcal{Q}$ and a gallery set $\mathcal{G}$, we find $N$ images from $\mathcal{G}$ that best match the images in $\mathcal{Q}$ in a query-balanced fashion, as explained in \Cref{sec:rats}. In our experiments, $\mathcal{Q}$ consists of a few images per class from the domain of interest. For example, in the case of EuroSAT with 5 images per class, $\mathcal{Q}$ consists of a total of $5 \times 10 = 50$ images. We performed both image-level and crop-level retrievals, as explained below.

\subsection{Image-level Retrieval from YFCC15M}
\label{sec:image-level-ret}
In this setup, we use the YFCC15M dataset~\cite{CLIP}, which contains 15M images, as the gallery set and the OpenCLIP's~\citep{Openclip} ViT-L/14 image encoder, trained on the DataComp-1B dataset~\citep{Datacomp}, as the encoder network. To obtain image-level features for both gallery and query sets, we resize and center-crop them to $224 \times 224$ and run the image encoder. We used this retrieval strategy to obtain retrieval-augmented transfer sets for the ADE20K~\citep{Ade20k} query sets.

\subsection{Crop-level Retrieval from DataComp}\label{sec:crop-level-ret}
In this setup, we use the DataComp-1B dataset~\cite{Datacomp}, which contains 1.28B (image, text) pairs (the best pool filtered of 12B images) as the gallery set. For each image in the gallery set, we consider 10 random crops using PyTorch's \verb|RandomResizeCrop| augmentation with scale parameters (0.08, 1.0) and a final crop size of $224 \times 224$. We further apply RandAug~\citep{RandAug} to each crop to increase the diversity of gallery crops. Our cropping and augmentation strategy results in a rich and diverse set of 12.8B gallery crops, allowing us to retrieve target task related crops from seemingly out-of-domain images (see~\cref{fig:eurosat_datacomp} in~\Cref{sec:ret_visual}).

For each crop in the gallery, we store two normalized features obtained by OpenCLIP's~\citep{Openclip} ViT-L/14 image encoders trained with OpenAI and Datacomp-1B pretraining datasets. For efficient retrieval, we use the cached feature store as in the dataset reinforcement strategy of \citet{MobileCLIP}. For query images, we resize and center-crop to $224 \times 224$ and run the same image encoders. For a pair of a gallery crop and a query image, we compute their similarity (for the $k$-NN search described in \Cref{sec:rats}) by averaging the cosine similarity of their features computed by the two image encoders. We used this retrieval strategy to obtain retrieval-augmented transfer sets for the EuroSAT~\citep{EuroSAT} query sets.

\subsection{De-duplication} \label{sec:dedup}
Web-scale datasets like DataComp, which we use to generate retrieval-augmented transfer sets, may include duplicated images. We found that the original de-duplication performed in DataComp~\citep{Datacomp} is not sufficient to remove all duplicates. To enhance the quality and diversity of the retrieved sets, we implement a de-duplication process as follows.

For every pair of retrieved crops, we compute the features of their original images using the same two ViT-L/14 image encoders from OpenCLIP~\citep{Openclip}, after resizing and center-cropping to $224 \times 224$. If the average similarity between the original images of a pair exceeds 0.99, we consider this pair a duplicate. We then cluster all duplicate pairs and ensure that only one instance from each cluster is included in the final retrieved set.

\subsection{De-contamination} \label{sec:decontam}
In our experiments, we use only a subset of a target task dataset as the query set to perform retrieval. It is possible that some other images from the target task dataset might have "leaked" into our gallery set (i.e., DataComp). To ensure we are not retrieving such images, we follow a de-contamination process as described below.

First, we form a set $\mathcal{S}$ of all images present in a given target task dataset (including train, val and test splits). We then compute two image-level features for all images in $\mathcal{S}$ using the same two ViT-L/14 image encoders from OpenCLIP~\citep{Openclip}, after resizing and center-cropping to $224 \times 224$. Similarly, we compute two image-level features for the original images of the retrieved crops. For every pair of images $(x_s, x_r)$, where $x_s \in \mathcal{S}$ and $x_r$ is the original image of a retrieved crop, we examine the maximum similarity between their two features. If the maximum similarity exceeds 0.95, we mark $x_r$ as a "possible leak". Next, for every $x_r$ identified as a "possible leak", we find its 5 nearest neighbors in $\mathcal{S}$ and perform a human visual inspection to verify whether it is an actual leak. Finally, we remove all retrieved crops corresponding to a leaked image.

\subsection{Retrieval Ablations}
\label{sec:retrieval_ablations}
Let $\mathcal{Q}$ denote the query set, $\mathcal{G}$ denote the gallery set and $\phi$ denote the encoder network. Initially, we evaluated the transfer sets created with the following four retrieval strategies under task-agnostic knowledge transfer setting. 
\begin{itemize}
    \item \textbf{Random:} Randomly select images from the gallery.
    \item  \textbf{Best-matches}: For each image $x \in \mathcal{G}$, we use $\min_{x_q \in \mathcal{Q}} {\| \phi(x) - \phi(x_q) \|_2}$ as its distance to the query set $\mathcal{Q}$. We retrieve images from $\mathcal{G}$ in the increasing order of their distance to $\mathcal{Q}$. 
    \item \textbf{Query-balanced (Image)}: For a query image $x_q \in \mathcal{Q}$, we define $k$-NN$(x_q)$ to be the set of its $k$ nearest neighbors from the gallery $\mathcal{G}$. To retrieve $N$ images in total, we find the smallest $k$ for which $\bigcup_{x_q \in \mathcal{Q}}k$-NN$(x_q)$ contains at least $N$ images. If $\bigcup_{x_q \in \mathcal{Q}}k$-NN$(x_q)$ contains more than $N$ images, we drop the $k^{th}$ neighbor of randomly selected queries until the retrieved set contains $N$ images.
    \item \textbf{Query-balanced (Text):} First, we convert the class names in the target task into text descriptions using the templates from~\citep{CLIP} and encode these text descriptions using the text encoder of the OpenCLIP model used to encode images. Then, we follow the above query balanced retrieval strategy using text queries instead of image queries. 
\end{itemize}

\begin{figure*}[!ht]
\centering
\includegraphics[scale=0.4, bb=0 0 550 500]{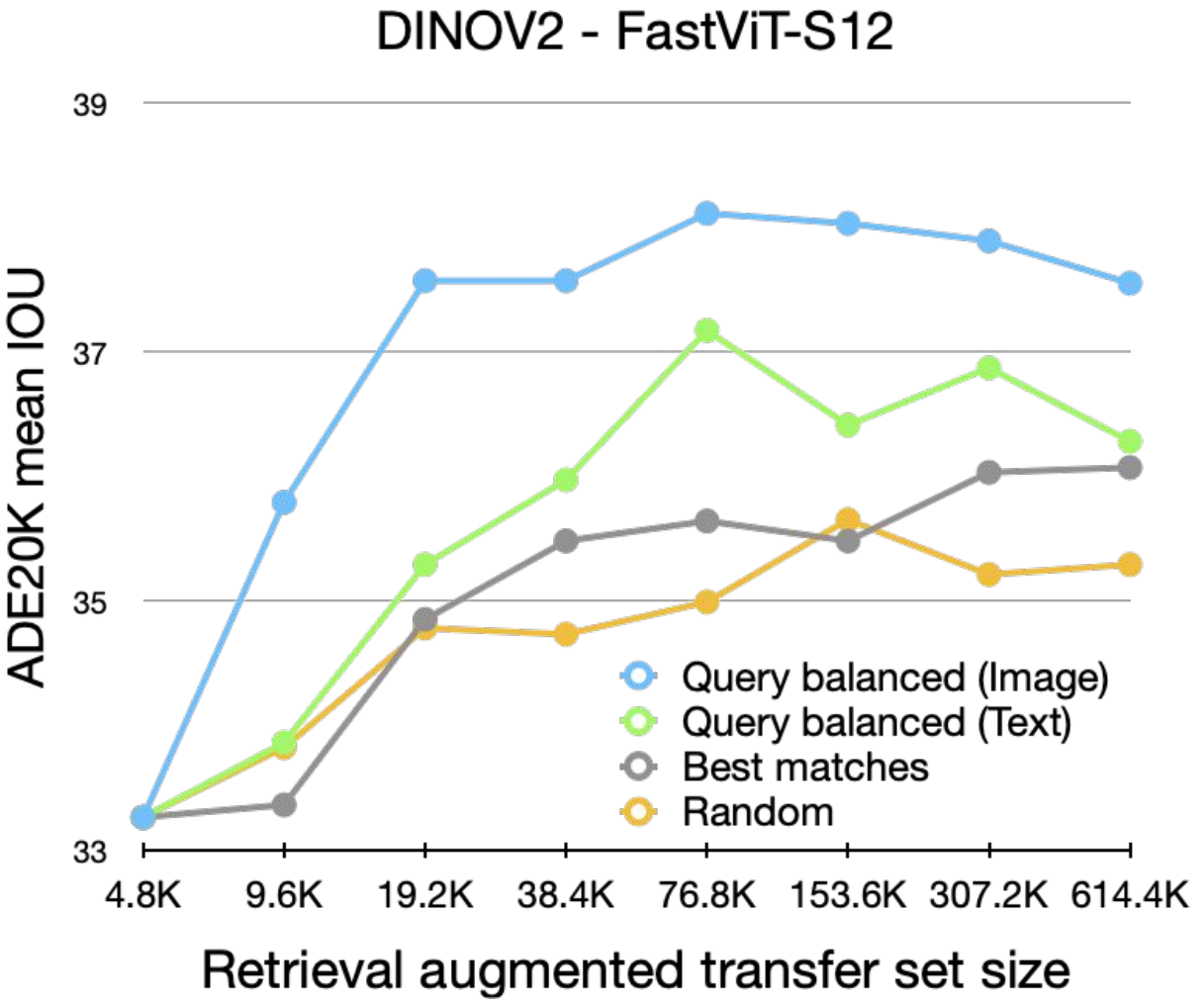}
\caption{Performance of task-agnostic transfer using transfer sets curated by different retrieval strategies. Here, we use 4800 labeled ADE20K images both as the finetuning dataset and the query set.}
\label{fig:retrieval_methods_ablations}
\end{figure*}

\Cref{fig:retrieval_methods_ablations} shows ADE20K segmentation performance for task-agnostic knowledge transfer using transfer sets curated by different retrieval strategies. Here, we use 4800 labeled images for finetuning the target model and use the same 4800 images as the query set for retrieval. Query-balanced retrieval based on image queries performs the best. By giving equal weight to all queries, this approach increases diversity in the retrieved samples when compared to the best-matches strategy. Following these initial results, we adopted the query-balanced retrieval strategy for all our experiments.

\subsection{Retrieval Visualizations} \label{sec:ret_visual}
\begin{figure}[!ht]
    \centering
    \includegraphics[scale=1.3, bb=0 0 365 420]{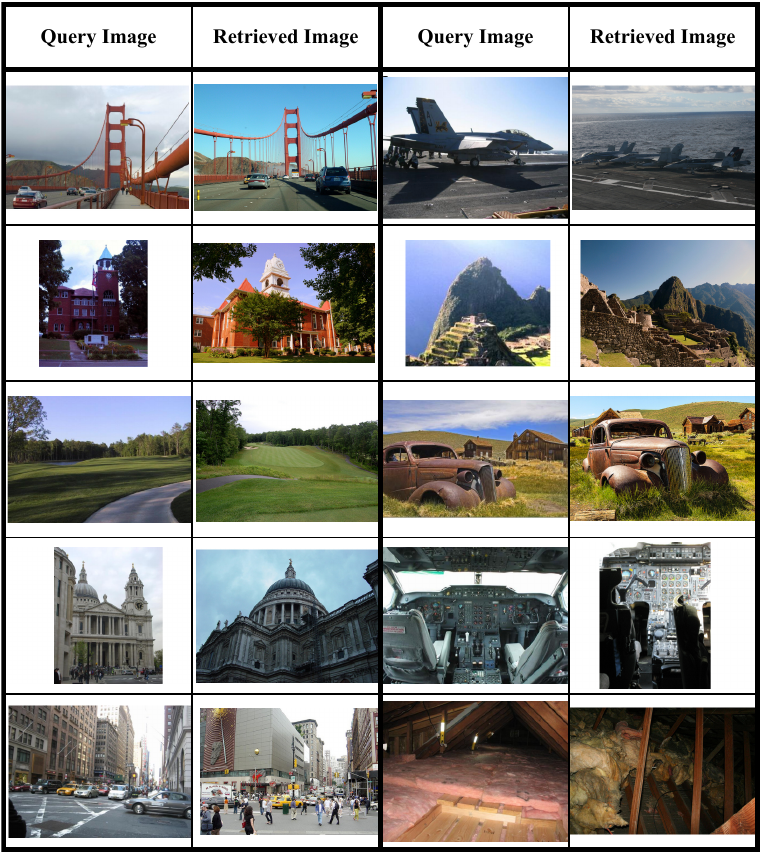}
    \caption{Pairs of ADE20K images (query) and their matched retrieved images from the YFCC15M dataset using the image-level retrieval algorithm  presented in \Cref{sec:image-level-ret}. Since YFCC15M and ADE20K come from similar distributions, image-level retrieval works well.}
    \label{fig:ade20_yfcc15m}
\end{figure}

\begin{figure}[!ht]
    \centering
    \includegraphics[scale=1.3, bb= 0 0 365 430]{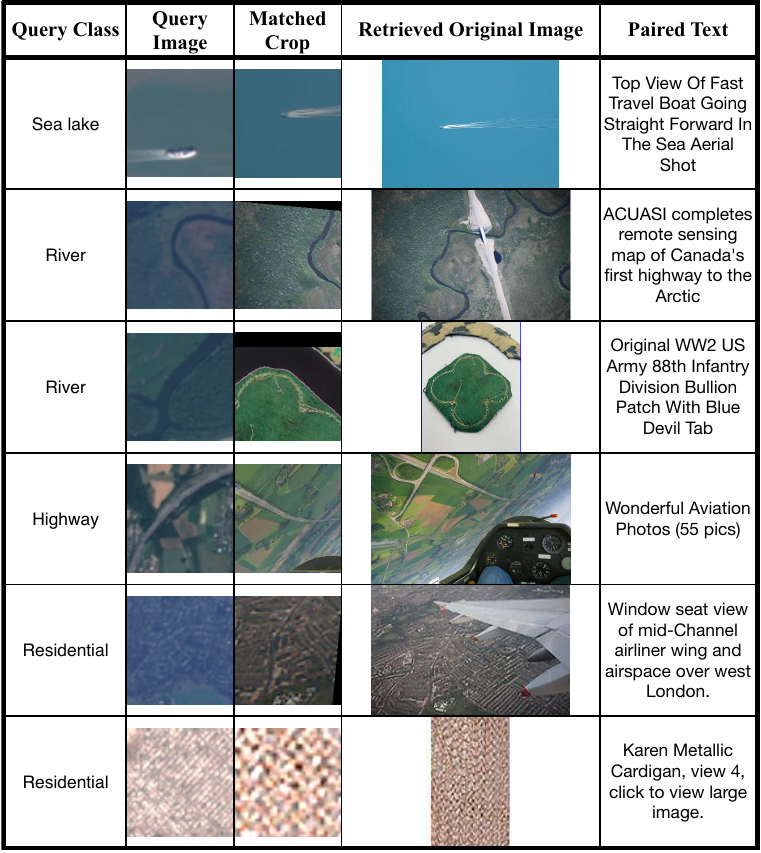}
    \caption{Pairs of EuroSAT images (query) and their matched retrieved images from the DataComp-1B dataset using a crop-level retrieval algorithm presented in \Cref{sec:crop-level-ret}. For each image, we show the matching crop and the original image, as well as the paired text from the gallery set. With our proposed crop-based retrieval, we can extract only in-domain parts of images (e.g., rows 2, 4, and 5). Furthermore, the combination of cropping and augmentation provides us with in-domain crops from unrelated images (e.g., rows 3 and 6).}
    \label{fig:eurosat_datacomp}
\end{figure}

In this section, we present some (query, retrieved) image pairs. \Cref{fig:ade20_yfcc15m} illustrates image-level retrieval from the YFCC15M dataset for queries from the ADE20K dataset, and \Cref{fig:eurosat_datacomp} shows crop-level retrieval from the DataComp-1B for queries from the EuroSAT dataset. Crop-level retrieval allows for patch extraction from a broader range of images, even those not directly related to the task.

\newpage
\begin{figure*}[ht]
\centering
\includegraphics[scale=0.68, bb=0 0 720 580]{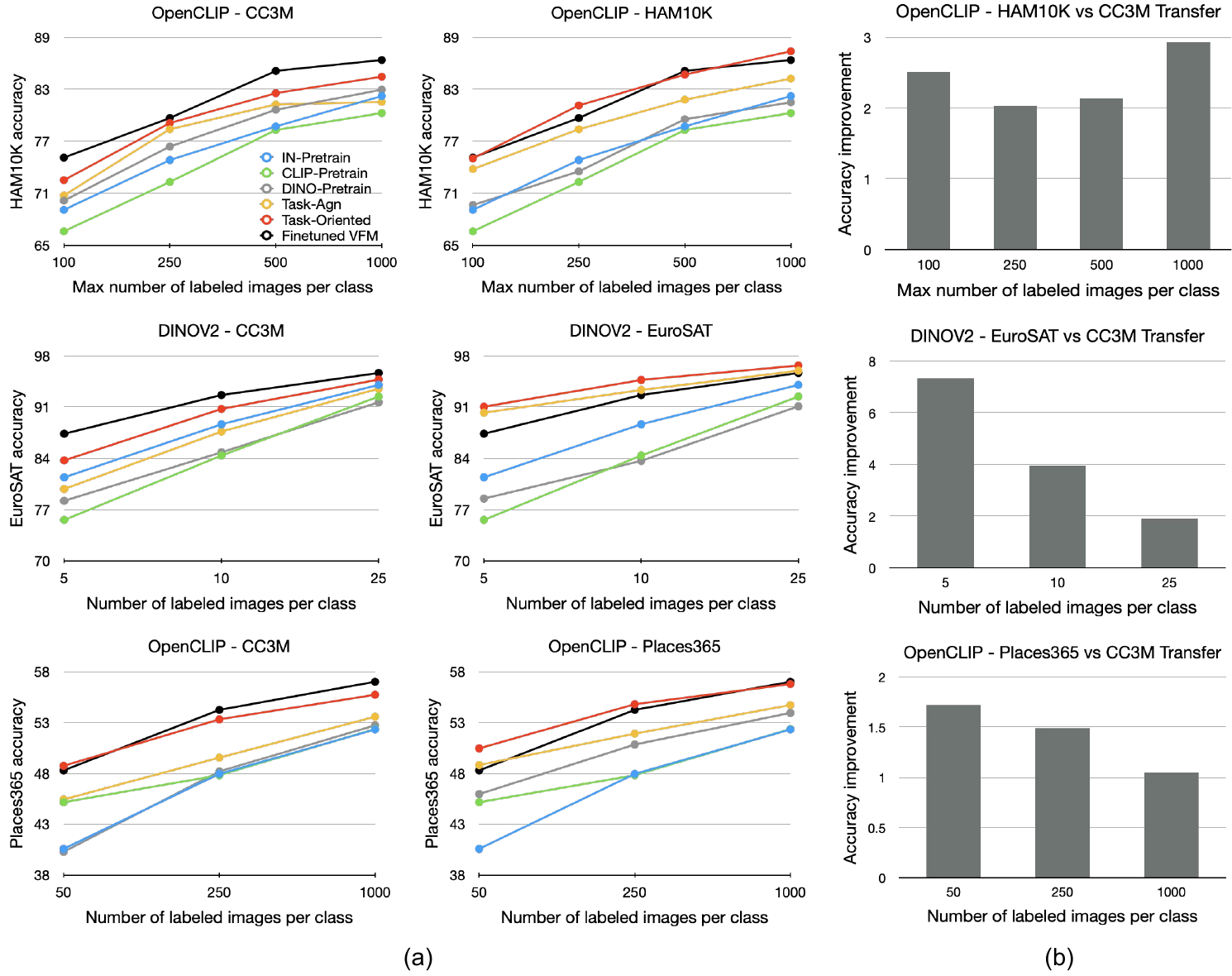}
\caption{(a) Comparison of various approaches for different VFM - Transfer set combinations with FastViT as the target image encoder. (b) Performance improvement when target task dataset is used instead of generic CC3M transfer set for task-oriented knowledge transfer. The target tasks are HAM10K, EuroSAT and Places365 classification from top to bottom. Task-oriented knowledge transfer from VFMs clearly outperforms various popular alternative training strategies.}
\label{fig:additional_fastvit_results}
\end{figure*} 
\section{Additional Results}
\subsection{FastViT Results}
\Cref{fig:main_results} in the main paper compares various approaches for different VFM and transfer set combinations on five downstream tasks with FastViT as the target image encoder.~\Cref{fig:additional_fastvit_results} in this Appendix presents additional results corresponding to few more combinations of VFMs and transfer sets. Task-oriented knowledge transfer from VFMs clearly outperforms various popular alternative training strategies.

\Cref{tab:ham10k_fastvit_results,tab:ade20k_fastvit_results,tab:imagenet_fastvit_results,tab:eurosat_fastvit_results,tab:places365_fastvit_results} present our experimental results on ImageNet, EuroSAT, Places365, ADE20K and HAM10K datasets, respectively, in tabular format. The values in these tables correspond to the plots in~\Cref{fig:main_results,fig:additional_fastvit_results}.
\label{sec:additional_fastvit_results}

\begin{table}[htb]
    \begin{scriptsize}
    \begin{center}
    \begin{tabular}{|llcccc|}
    \hline
         \multicolumn{2}{|l}{Percentage of labeled training images} & 1 & 10 & 25 & 50 \\\hline
         \multicolumn{2}{|l}{Finetuned VFM (OpenCLIP)} & $74.72^{*}$ & $83.31^{\hspace{3pt}}$ & $84.98$ & $86.05$\\\hline\hline
        \multicolumn{2}{|l}{From scratch} & $14.46^{\hspace{3pt}}$ & $56.35^{\hspace{3pt}}$ & $68.90$ & $75.17$\\\hline
        \multicolumn{2}{|l}{CLIP-Pretrain} & $51.94^{*}$ &	$65.00^{*}$ & $71.93$ & $76.05$\\\hline

        \multicolumn{2}{|l}{DINO-Pretrain (CC3M)} & 
        $33.29^{\hspace{3pt}}$ & $63.89^{\hspace{3pt}}$ & $71.36$ & $75.97$\\\hline
        \multicolumn{2}{|l}{DINO-Pretrain (ImageNet)} & $52.27^{*}$ & $70.64^{\hspace{3pt}}$ & $74.79$ & $77.41$ \\\hline

         \multirow{2}{*}{OpenCLIP} & \multicolumn{1}{|l}{Task-Agn} & 
         $51.47^{*}$ & $68.12^{\hspace{3pt}}$ & $73.24$ & $76.53$\\\cline{2-6}
         (CC3M transfer) & \multicolumn{1}{|l}{Task-Oriented} & 
         $63.13^{\hspace{3pt}}$ & $74.29^{\hspace{3pt}}$ & $76.79$ & $78.40$\\\hline

         \multirow{2}{*}{OpenCLIP} & \multicolumn{1}{|l}{Task-Agn} & 
         $67.74^{*}$ & $74.81^{*}$ & $76.01$ & $77.97$\\\cline{2-6}
         (ImageNet transfer) & \multicolumn{1}{|l}{Task-Oriented} & 
         $74.03^{\hspace{3pt}}$ & $80.08^{\hspace{3pt}}$ & $80.91$ & $81.43$\\\hline
    \end{tabular}
    \end{center}
    \end{scriptsize}
    \vspace{-5pt}
    \caption{ImageNet classification accuracy for FastViT target model. The results marked with $^*$ are obtained by training only the classification layer instead of the entire model in the finetuning stage. Full finetuning produced inferior results in these cases.}
    \label{tab:imagenet_fastvit_results}
\end{table}
\begin{table}[htb!]
    \begin{scriptsize}
    \begin{center}
    \begin{tabular}{|llccc|}
    \hline
         \multicolumn{2}{|l}{Labeled training images per class} & 5 & 10 & 25\\\hline
         \multicolumn{2}{|l}{Finetuned VFM (DINOv2)} & $87.36$ & $92.65$ & $95.66$ \\\hline
         \multicolumn{2}{|l}{Finetuned VFM (OpenCLIP)} & $87.99$ & $92.63$ & $96.08$\\\hline\hline
        
        \multicolumn{2}{|l}{IN-Pretrain} & $81.41$ & $88.65$ & $94.04$\\\hline
          \multicolumn{2}{|l}{CLIP-Pretrain} & $75.59$ & $84.38$ & $92.44$\\\hline
          
        \multicolumn{2}{|l}{DINO-Pretrain (CC3M)} & $78.20$ & $84.83$ & $91.66$\\\hline
        \multicolumn{2}{|l}{DINO-Pretrain (EuroSAT)} & $78.50$ & $83.65$ & $91.11$\\\hline
        
        \multirow{2}{*}{DINOv2} & \multicolumn{1}{|l}{Task-Agn (Patch)} & $76.52$ & $86.11$ & $92.66$\\\cline{2-5}
        (CC3M transfer) & \multicolumn{1}{|l}{Task-Agn (Image)} &
        $79.81$ &	$87.66$ & $93.52$\\\cline{2-5}
        & \multicolumn{1}{|l}{Task-Oriented} & $83.72$ & $90.75$ & $94.77$\\\hline
            
        \multirow{2}{*}{DINOv2} & \multicolumn{1}{|l}{Task-Agn (Patch)} & $90.23$ & $93.34$ & $95.97$\\\cline{2-5}        
        (EuroSAT transfer) & \multicolumn{1}{|l}{Task-Agn (Image)} & $86.54$ & $89.56$ & $94.19$\\\cline{2-5}
        & \multicolumn{1}{|l}{Task-Oriented} & $91.05$ & $94.71$ & $96.68$\\\hline
           
         \multirow{2}{*}{OpenCLIP} & \multicolumn{1}{|l}{Task-Agn} & $79.87$ & $87.10$ & $92.91$ \\\cline{2-5}
         (CC3M transfer) & \multicolumn{1}{|l}{Task-Oriented} & 
         $85.14$ & $90.21$ & $94.25$\\\hline

         \multirow{2}{*}{OpenCLIP} & \multicolumn{1}{|l}{Task-Agn} & $88.57$ &	$92.28$ & $95.5$\\\cline{2-5}
         (EuroSAT transfer) & \multicolumn{1}{|l}{Task-Oriented} & 
         $90.74$ & $94.63$ & $96.83$\\\hline

    \end{tabular}
    \end{center}
    \end{scriptsize}
    \vspace{-5pt}
    \caption{EuroSAT classification accuracy for FastViT target model.}
    \label{tab:eurosat_fastvit_results}
\end{table}

\begin{table}[htb!]
    \begin{scriptsize}
    \begin{center}
    \begin{tabular}{|llccc|}
    \hline
         \multicolumn{2}{|l}{Labeled training images per class} & 50 & 250 & 1000 \\\hline
         \multicolumn{2}{|l}{Finetuned VFM (DINOv2)} & $47.56 \pm 0.02^{\hspace{3pt}}$ & $54.11 \pm 0.27$ & $56.95 \pm 0.11$  \\\hline
         \multicolumn{2}{|l}{Finetuned VFM (OpenCLIP)} & $48.30 \pm 0.59^{\hspace{3pt}}$ & $54.26 \pm 0.19$ & $57.03 \pm 0.17$ \\\hline\hline
        \multicolumn{2}{|l}{IN-Pretrain} &  $40.58 \pm 0.14^{\hspace{3pt}}$ & $47.96 \pm 0.09$ & $52.33 \pm 0.14$\\\hline
          \multicolumn{2}{|l}{CLIP-Pretrain} & $45.17 \pm 0.03^*$ & $47.83 \pm 0.30$ & $52.37 \pm 0.47$ \\\hline
          \multicolumn{2}{|l}{DINO-Pretrain (CC3M)}
          & $40.29 \pm 0.26^{\hspace{3pt}}$ & $48.20 \pm 0.16$ & $52.74 \pm 0.12$\\\hline
           \multicolumn{2}{|l}{DINO-Pretrain (Places365)} & 
           $45.97 \pm 0.18^{\hspace{3pt}}$ & $50.85 \pm 0.12$ & $53.96 \pm  0.06$ \\\hline 

         \multirow{2}{*}{DINOv2} & \multicolumn{1}{|l}{Task-Agn (Patch)} & $42.46 \pm 0.02^*$ & $49.60 \pm 0.11$ & $53.49 \pm 0.06$ \\\cline{2-5}
         (CC3M transfer) & \multicolumn{1}{|l}{Task-Agn (Image)} & $44.52 \pm 0.13^*$ &  $49.75 \pm 0.07$ & $53.45 \pm 0.04$ \\\cline{2-5}
           & \multicolumn{1}{|l}{Task-Oriented} & $47.81 \pm 0.05^{\hspace{3pt}}$ & $53.09 \pm 0.05$ & $55.62 \pm 0.05$ \\\hline
          
          \multirow{2}{*}{DINOv2} & \multicolumn{1}{|l}{Task-Agn (Patch)} & $46.45 \pm 0.05^*$ &  $51.26 \pm 0.20$ & $54.43 \pm 0.03$\\\cline{2-5}
          (Places365 transfer) & \multicolumn{1}{|l}{Task-Agn (Image)} & $47.76 \pm 0.04^*$ &  $51.45 \pm 0.32$ & $54.45 \pm 0.21$\\\cline{2-5}
          & \multicolumn{1}{|l}{Task-Oriented} & $49.14 \pm 0.02^{\hspace{3pt}}$ & $54.51 \pm 0.05$ & $56.68 \pm 0.01$ \\\hline
          
          \multirow{2}{*}{OpenCLIP} & \multicolumn{1}{|l}{Task-Agn} & $45.44 \pm  0.03^*$ & $49.56 \pm 0.05$ & $53.59 \pm 0.09$ \\\cline{2-5}
          (CC3M transfer) & \multicolumn{1}{|l}{Task-Oriented} & $48.75 \pm 0.01^{\hspace{3pt}}$ &	$53.33 \pm 0.05$ & $55.75 \pm 0.06$ \\\hline
          
           \multirow{2}{*}{OpenCLIP} & \multicolumn{1}{|l}{Task-Agn} & $48.83 \pm 0.07^*$ & $51.92 \pm 0.26$ & $54.74 \pm 0.10$ \\\cline{2-5}
           (Places365 transfer) & \multicolumn{1}{|l}{Task-Oriented} & $50.47 \pm 0.01^{\hspace{3pt}}$ & $54.82 \pm 0.04$ & $56.80 \pm 0.02$\\\hline
    \end{tabular}
    \end{center}
    \end{scriptsize}
    \vspace{-5pt}
    \caption{Places365 classification accuracy for FastViT target model. The results marked with $^*$ are obtained by training only the classification layer instead of the entire model in the finetuning stage. Full finetuning produced inferior results in these cases.}
    \label{tab:places365_fastvit_results}
\end{table}

\begin{table}[htb!]
    \begin{scriptsize}
    \begin{center}
    \begin{tabular}{|llcccc|}
    \hline
         \multicolumn{2}{|l}{Labeled training images} & 1200 & 2401 & 4802 & 9605 \\\hline
         \multicolumn{2}{|l}{Finetuned VFM (DINOv2)} & $39.32 \pm 0.11$ & $42.76 \pm 0.08$ & $46.35 \pm 0.09$ & $49.42 \pm 0.14$ \\\hline\hline
        \multicolumn{2}{|l}{IN-Pretrain} & $22.58 \pm 0.07$ & $27.11 \pm 0.06$ & $33.12 \pm 0.25$ & $37.54 \pm 0.02$\\\hline
          \multicolumn{2}{|l}{CLIP-Pretrain}& $15.34 \pm 0.20$ & $20.88 \pm 0.26$ & $27.57 \pm 0.11$ & $32.63 \pm 0.08$ \\\hline
          
          \multicolumn{2}{|l}{DINO-Pretrain (CC3M)} & 
          $20.88 \pm 0.27$ & $25.75 \pm 0.25$ &	$31.74 \pm 0.08$ & $36.43 \pm 0.02$\\\hline
          \multicolumn{2}{|l}{DINO-Pretrain (ADE20K)} & 
          $16.43 \pm 0.09$ & $21.14 \pm 0.20$ & $27.71 \pm 0.23$ & $32.23 \pm 0.23$\\\hline

         \multirow{2}{*}{DINOv2} & \multicolumn{1}{|l}{Task-Agn (Patch)} & $25.75 \pm 0.55$ & $30.27 \pm 0.14$ & $35.83 \pm 0.11$ & $39.28 \pm 0.54$\\\cline{2-6}
          (CC3M transfer) & \multicolumn{1}{|l}{Task-Oriented} &
           $36.28 \pm 0.01$	& $39.22 \pm 0.19$ & $41.44 \pm 0.15$ & $44.29 \pm 0.03$ \\\hline

        \multirow{2}{*}{DINOv2} & \multicolumn{1}{|l}{Task-Agn (Patch)}  & $28.32 \pm 0.46$ & $32.0 \pm 0.19$ & $36.81\pm 0.08$ &  $39.44 \pm 0.22$ \\\cline{2-6}
          (ADE20K transfer) & \multicolumn{1}{|l}{Task-Oriented} & $34.57 \pm 0.02$ & $37.19 \pm 0.10$ & $38.45 \pm 0.05$ & $41.07 \pm 0.21$ \\\hline
    \end{tabular}
    \end{center}
    \end{scriptsize}
    \vspace{-5pt}
    \caption{ADE20K mean IOU for FastViT target model.}
    \label{tab:ade20k_fastvit_results}
\end{table}

\begin{table}[htb]
    \begin{scriptsize}
    \begin{center}
    \begin{tabular}{|llcccc|}
    \hline
         \multicolumn{2}{|l}{Maximum labeled training images per class} & 100 & 250 & 500 & 1000 \\\hline
         \multicolumn{2}{|l}{Finetuned VFM (DINOv2)} & 
         $77.78 \pm 0.61$ &  $81.79 \pm 0.97$ & $84.57 \pm 1.88$ & $88.12 \pm 0.56$ \\\hline
         \multicolumn{2}{|l}{Finetuned VFM (OpenCLIP)}  & 
         $75.13 \pm 0.82$ & $79.70 \pm 1.19$ & $85.12 \pm 0.35$ & $86.38 \pm 0.74$\\\hline\hline
        \multicolumn{2}{|l}{IN-Pretrain} & $69.11 \pm 1.56$ & $74.85 \pm 0.82$ & $78.73 \pm 1.88$ & $82.23 \pm 1.37$ \\\hline
          \multicolumn{2}{|l}{CLIP-Pretrain} & $66.64 \pm 1.75$ & $72.33 \pm 1.11$ & $78.31 \pm 1.99$ & $80.27 \pm 2.08$ \\\hline
          \multicolumn{2}{|l}{DINO-Pretrain (CC3M)} &
         	$70.17 \pm 1.60$ & $76.41 \pm 1.00$ & $80.64 \pm 0.76$ & $82.96 \pm 0.31$\\\hline
          \multicolumn{2}{|l}{DINO-Pretrain (HAM10K)} & $69.66 \pm 2.52$ & $73.54 \pm 1.86$ & $79.56 \pm 0.81$ & $81.50 \pm 1.46$\\\hline

         \multirow{2}{*}{DINOv2} & \multicolumn{1}{|l}{Task-Agn (Patch)} & $71.76 \pm 2.28$ & $75.68 \pm 1.83$ & $80.34 \pm 0.74$ & $80.95 \pm 1.16$ \\\cline{2-6}
          (CC3M transfer) & \multicolumn{1}{|l}{Task-Agn (Image)} & $71.32 \pm 1.94$ & $77.67 \pm 1.48$ & $80.58 \pm 0.53$ & $82.78 \pm 1.01$  \\\cline{2-6}
           & \multicolumn{1}{|l}{Task-Oriented} & $73.06 \pm 0.79$ & $78.44 \pm 0.41$ & $81.88 \pm 0.64$ & $83.16 \pm 0.83$\\\hline
            
         \multirow{2}{*}{DINOv2} & \multicolumn{1}{|l}{Task-Agn (Patch)} & $75.29 \pm 1.22$ & $78.77 \pm 1.55$ & $83.96 \pm 0.10$ & $85.19 \pm 0.40$  \\\cline{2-6}         
          (HAM10K transfer) & \multicolumn{1}{|l}{Task-Agn (Image)} & $72.18 \pm 2.57$ & $76.85 \pm 0.74$ & $81.68 \pm 0.85$ & $83.64 \pm 0.59$ \\\cline{2-6}
           & \multicolumn{1}{|l}{Task-Oriented} & $78.04 \pm 0.05$ & $82.30 \pm 0.14$ & $86.33 \pm 0.16$ & $86.20 \pm 0.32$\\\hline
           
         \multirow{2}{*}{OpenCLIP} & \multicolumn{1}{|l}{Task-Agn} & $70.77 \pm 1.00$ & $78.40 \pm 0.77$ & $81.26 \pm 0.37$ & $81.55 \pm 1.31$ \\\cline{2-6}
         (CC3M transfer) & \multicolumn{1}{|l}{Task-Oriented} & $72.53 \pm 0.63$ & 
         $79.12 \pm 1.27$ & $82.56 \pm 0.4$ & $84.46 \pm 0.65$\\\hline

         \multirow{2}{*}{OpenCLIP} & \multicolumn{1}{|l}{Task-Agn} & $73.81 \pm 0.44$ & $78.40 \pm 0.27$ & $81.81 \pm 0.82$ & $84.24 \pm 1.53$ \\\cline{2-6}
         (HAM10K transfer) & \multicolumn{1}{|l}{Task-Oriented} & $75.04 \pm 0.06$ & $81.15 \pm 0.01$ & $84.70 \pm 0.08$ & $87.39 \pm 0.08$\\\hline
    \end{tabular}
    \end{center}
    \end{scriptsize}
    \vspace{-5pt}
    \caption{HAM10K classification accuracy for FastViT target model.}
    \label{tab:ham10k_fastvit_results}
\end{table}

\newpage
\subsection{MViT-V2 Results}
\label{sec:MVIT_results}
\begin{figure*}[b!]
\centering
\includegraphics[scale=0.65,bb=200 0 520 500]{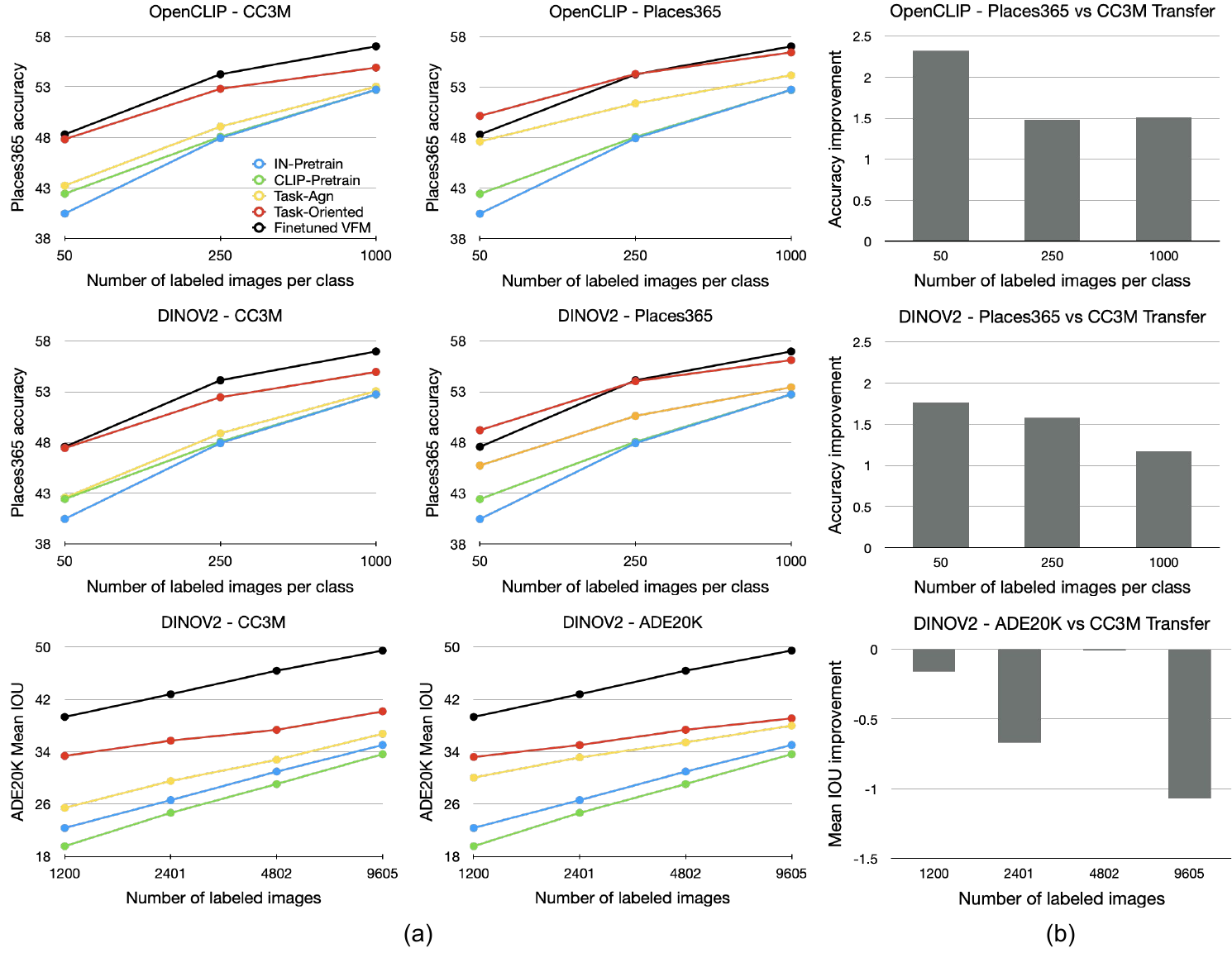}
\vspace{-5pt}
\caption{(a) Comparison of various approaches for different VFM - Transfer set combinations with MViT-V2 as the target image encoder. (b) Performance improvement when target task dataset is used instead of generic CC3M transfer set for task-oriented knowledge transfer. The target tasks are Places365 classification and ADE20K segmentation from top to bottom. Task-oriented knowledge transfer from VFMs clearly outperforms various popular alternative training strategies.}
\label{fig:mvit_results}
\end{figure*}

The experimental results with FastViT as the target image encoder are presented in the main paper and~\Cref{sec:additional_fastvit_results}.~\Cref{fig:mvit_results} in this Appendix presents results for some VFM and transfer set combinations with MViT-V2 as the target image encoder. Task-oriented knowledge transfer from VFMs clearly outperforms various popular alternative training strategies. For CLIP pretraining of MViT-V2 model, we use the original CLIP loss from~\citep{CLIP} instead of the affinity mimicking distillation loss of~\cite{TinyCLIP}. 

\Cref{tab:places365_mvit_results,tab:ade20k_mvit_results} present our experimental results on Places365 and ADE20K datasets, respectively, in tabular format. The values in these tables correspond to the plots in~\Cref{fig:mvit_results}.

\begin{table}[h!]
    \begin{scriptsize}
    \begin{center}
    \begin{tabular}{|llcccc|}
    \hline
         \multicolumn{2}{|l}{Labeled training images} & 1200 & 2401 & 4802 & 9605 \\\hline
         \multicolumn{2}{|l}{Finetuned VFM (DINOv2)} & $39.32 \pm 0.11$ & $42.76 \pm 0.08$ & $46.35 \pm 0.09$ & $49.42 \pm 0.14$ \\\hline\hline
          \multicolumn{2}{|l}{IN-Pretrain} & $22.36 \pm 0.18$ & $26.60 \pm 0.26$ & $30.95 \pm 0.41$ & $35.02 \pm 0.01$ \\\hline
          \multicolumn{2}{|l}{CLIP-Pretrain} & $19.58 \pm 0.22$ & $24.66 \pm 0.09$ & $29.06 \pm 0.24$ & $33.62 \pm 0.14$ \\\hline
          
         \multirow{2}{*}{DINOv2} & \multicolumn{1}{|l}{Task-Agn (Patch)} & $25.42 \pm 0.18$ & $29.52 \pm 0.10$ & $32.76 \pm 0.28$ & $36.73 \pm 0.17$ \\\cline{2-6}
          (CC3M transfer) & \multicolumn{1}{|l}{Task-Oriented} & 
         $33.35 \pm 0.02$ & $35.68 \pm 0.08$ & $37.32 \pm 0.11$ & $40.14 \pm 0.05$ \\\hline

        \multirow{2}{*}{DINOv2} & \multicolumn{1}{|l}{Task-Agn (Patch)} & $30.03 \pm 0.15$ & $33.12 \pm 0.27$ & $35.41 \pm 0.15$ & $37.96 \pm 0.39$ \\\cline{2-6}
          (ADE20K transfer) & \multicolumn{1}{|l}{Task-Oriented} & $33.19 \pm 0.11$ & $35.01 \pm 0.09$ & $37.33 \pm 0.03$ & $39.07 \pm 0.09$ \\\hline
    \end{tabular}
    \end{center}
    \end{scriptsize}
    \vspace{-5pt}
    \caption{ADE20K mean IOU for MViT-V2 target model.}
    \label{tab:ade20k_mvit_results}
\end{table}

\begin{table}[ht!]
    \begin{scriptsize}
    \begin{center}
    \begin{tabular}{|llccc|}
    \hline
         \multicolumn{2}{|l}{Labeled images per class} & 50 & 250 & 1000 \\\hline
         \multicolumn{2}{|l}{Finetuned VFM (DINOv2)} & $47.56 \pm 0.02^{\hspace{3pt}}$ & $54.11 \pm 0.27$ & $56.95 \pm 0.11$  \\\hline
         \multicolumn{2}{|l}{Finetuned VFM (OpenCLIP)} & $48.30 \pm 0.59^{\hspace{3pt}}$ & $54.26 \pm 0.19$ & $57.03 \pm 0.17$ \\\hline\hline
          \multicolumn{2}{|l}{IN-Pretrain} & $40.46 \pm 0.07^{\hspace{3pt}}$ & $47.93 \pm 0.15$ & $52.73 \pm 0.04$ \\\hline
          \multicolumn{2}{|l}{CLIP-Pretrain} & $42.42 \pm 0.05^*$ & $48.06 \pm 0.16$ & $52.72 \pm 0.08$ \\\hline
          
         \multirow{2}{*}{DINOv2} & \multicolumn{1}{|l}{Task-Agn (Patch)} & $41.81 \pm 0.16^{\hspace{3pt}}$ & $48.90 \pm 0.12$ & $53.15 \pm 0.08$\\\cline{2-5}
           (CC3M transfer ) & \multicolumn{1}{|l}{Task-Agn (Image)} & $42.55 \pm 0.19^{\hspace{3pt}}$ & $48.89 \pm 0.09$ & $53.05 \pm 0.09$\\\cline{2-5}
            & \multicolumn{1}{|l}{Task-Oriented} & $47.44 \pm 0.02^{\hspace{3pt}}$ & $52.44 \pm 0.04$ & $54.93 \pm 0.04$ \\\hline
            
           \multirow{2}{*}{DINOv2} & \multicolumn{1}{|l}{Task-Agn (Patch)} & $45.14 \pm 0.08^{\hspace{3pt}}$ &	$50.37 \pm 0.06$ & $53.72 \pm 0.08$\\\cline{2-5}
            (Places365 transfer) & \multicolumn{1}{|l}{Task-Agn (Image)} & $45.73 \pm 0.02^{\hspace{3pt}}$ & $50.61 \pm 0.05$ & $53.43 \pm 0.03$\\\cline{2-5}
            & \multicolumn{1}{|l}{Task-Oriented} & $49.20 \pm 0.04^{\hspace{3pt}}$ & $54.02 \pm 0.05$ & $56.10 \pm 0.03$ \\\hline
            
          \multirow{2}{*}{OpenCLIP} & \multicolumn{1}{|l}{Task-Agn} &  $43.23 \pm 0.09^*$ & $49.07 \pm 0.04$ & $53.04 \pm 0.14$\\\cline{2-5}
         (CC3M transfer) & \multicolumn{1}{|l}{Task-Oriented} & $47.82 \pm 0.02^{\hspace{3pt}}$ & $52.82 \pm 0.05$ & $54.92 \pm 0.03$\\\hline
         
         \multirow{2}{*}{OpenCLIP} & \multicolumn{1}{|l}{Task-Agn} & $47.61 \pm 0.05^*$ & $51.39 \pm 0.10$ & $54.17 \pm 0.11$\\\cline{2-5}
         (Places365 transfer) & \multicolumn{1}{|l}{Task-Oriented} & $50.14 \pm 0.04^{\hspace{3pt}}$ & $54.30 \pm 0.02$ & $56.43 \pm 0.04$ \\\hline
    \end{tabular}
    \end{center}
    \end{scriptsize}
    \vspace{-5pt}
    \caption{Places365 classification accuracy for MViT-V2 target model. The results marked with $^*$ are obtained by training only the classification layer instead of the entire model in the finetuning stage. Full finetuning produced inferior results in these cases.}
    \label{tab:places365_mvit_results}
\end{table}

\subsection{Comparison with Pseudo-labeling based Semi-supervised Learning}
\Cref{tab:semi_supervised_results} shows the performance of a pseudo labeling-based semi-supervised approach with multiple rounds of training. In the first round, we obtain an initial model by training it with labeled data. In the subsequent rounds, we use the current model to generate pseudo labels for the unlabeled data and train a new model using both labeled and pseudo labeled data. In each round, the model is initialized with ImageNet pretrained checkpoint in the case of EuroSAT dataset and DINO(CC3M)-pretrained checkpoint in the case of HAM10K dataset. We use ImageNet pretraining for EuroSAT and DINO pretraining for HAM10K since these are the best performing baseline pretraining strategies for these datasets. In the case of EuroSAT classification, the performance increases when pseudo labeled data is used for the first time (round 2). However, the improvement is not significant in the next round. The final performance (87.8\%) after three rounds of training is still significantly lower than 91\% achieved by the proposed approach. In the case of HAM10K dataset, using pseudo labeled data results in negligible performance gain, and the proposed approach performs significantly better.

\begin{table}[t!]
    \begin{scriptsize}
    \begin{center}
    \begin{tabular}{|l|c|ccc|c|}
    \hline
         \multirow{2}{*}{Dataset} & \multirow{2}{*}{Labeled images per class} & \multicolumn{3}{c|}{Semi-supervised learning} & \multirow{2}{*}{Proposed} \\\cline{3-5}
        &  & Round 1 & Round 2 & Round 3 & \\\hline
         EuroSAT & 5 & 	81.4 & 87.2	& 87.8 & 91.0\\\hline
         HAM10K & 500 & 81.6 & 82.4 & - & 86.3\\\hline
    \end{tabular}
    \end{center}
    \end{scriptsize}
    \vspace{-5pt}
    \caption{Comparison with pseudo labeling-based semi-supervised learning using FastViT target model.}
    \label{tab:semi_supervised_results}
\end{table}

\subsection{Performance with Large Amount of Labeled Training Data}
The main focus of this work is on improving the target task performance with limited amount of labeled data. Hence, most of our experiments focus on limited labeled data settings. In this section, we compare the performance of different approaches when large amount of target task labeled data is available.

\Cref{tab:full_dataset_results} shows the performance achieved by various approaches on ImageNet and Places365 classification tasks when the full target task training dataset is used for both knowledge transfer and supervised finetuning (1.28M images in the case of ImageNet and 1.8M images in the case of Places365). We use FastViT as the target model for these experiments. The proposed task-oriented knowledge transfer approach outperforms all the other pretraining strategies by a significant margin in the case of ImageNet classification, and outperforms ImageNet/CLIP pretraining approaches by a significant margin in the case of Places365 classification.

\begin{table}[t!]
    \begin{scriptsize}
    \begin{center}
    \begin{tabular}{|l|c|cccccc|}
    \hline
    Dataset & VFM & From scratch & IN-Pretrain & CLIP-Pretrain & DINO-Pretrain & Task-Agn & Task-Oriented\\\hline
    ImageNet & OpenCLIP & 79.45 & - & 79.35 & 80.04 & 79.70 & 81.50\\\hline
    Places365 & DINOv2 & - & 56.74 & 56.59 & 57.15 & 57.51 & 57.54\\\hline
    \end{tabular}
    \end{center}
    \end{scriptsize}
    \vspace{-5pt}
    \caption{Performance of FastViT target model trained with various approaches under 100\% labeled training data setting.}
    \label{tab:full_dataset_results}
\end{table}

\end{document}